\documentclass{article}

\PassOptionsToPackage{numbers, compress}{natbib}

 \usepackage[preprint]{neurips_2026}

\usepackage[utf8]{inputenc} 
\usepackage[T1]{fontenc}    
\usepackage{hyperref}       
\usepackage{url}            
\usepackage{booktabs}       
\usepackage{amsfonts}       
\usepackage{nicefrac}       
\usepackage{microtype}      
\usepackage{xcolor}         

\usepackage{graphicx}
\usepackage{amsmath}
\usepackage{amssymb}
\usepackage{xspace}
\usepackage{tabularx}
\usepackage{multirow}
\usepackage{algorithm}
\usepackage{algorithmic}
\usepackage{bbding}
\usepackage{fontawesome5}

\newcommand{\sysname}{{\tt Motion-MLLM}\xspace}
\newcommand{\blue}[1]{\textcolor{black}{#1}}

\title{Feeling the Space: Egomotion-Aware Video Representation for Efficient and Accurate 3D Scene Understanding}

\author{%
  Shuyao Shi \\
  Department of Computer Science\\
  University of Michigan\\
  Ann Arbor, MI, USA \\
  \texttt{syshi@umich.edu} \\
  \And
  Kang G. Shin\thanks{Corresponding author.} \\
  Department of Computer Science\\
  University of Michigan\\
  Ann Arbor, MI, USA \\
  \texttt{kgshin@umich.edu} \\
}

\begin{document}

\maketitle
\vspace*{-0.1in}

\begin{abstract}
Recent Multimodal Large Language Models (MLLMs) have shown high potential for spatial reasoning within 3D scenes. However, they typically rely on computationally expensive 3D representations like point clouds or reconstructed Bird's-Eye View (BEV) maps, or lack physical grounding to resolve ambiguities in scale and size. This paper significantly enhances MLLMs with egomotion modality data, captured by Inertial Measurement Units (IMUs) concurrently with the video. In particular, we propose a novel framework, called {\tt Motion-MLLM}, introducing two key components: (1) a cascaded motion-visual keyframe filtering module that leverages both IMU data and visual features to efficiently select a sparse yet representative set of keyframes, and (2) an asymmetric cross-modal fusion module where motion tokens serve as intermediaries that channel egomotion cues and cross-frame visual context into the visual representation. By grounding visual content in physical egomotion trajectories, {\tt Motion-MLLM} can reason about absolute scale and spatial relationships across the scene. Our extensive evaluation shows that {\tt Motion-MLLM} makes significant improvements in various tasks related to 3D scene understanding and spatial reasoning. Compared to state-of-the-art (SOTA) methods based on video frames and explicit 3D data, {\tt Motion-MLLM} \blue{achieves competitive accuracy while running $1.30\times$ and $1.61\times$ faster, respectively}.
\end{abstract}

\section{Introduction}

Rapid advances in Multimodal Large Language Models (MLLMs)~\cite{achiam2023gpt, bai2025qwen25vltechnicalreport, team2024gemini, liu2023visual, li2023blip, alayrac2022flamingo} 
have demonstrated strong capabilities in jointly reasoning over 
multimodal inputs such as images, videos, and audio to produce 
contextually grounded responses~\cite{huang2024audiogpt, qian2024streaming, lin2024video, li2024llava}. 
This progress has led to the adoption of MLLMs in 
applications such as embodied AI~\cite{zheng2024towards, li2024manipllm}, robotic navigation~\cite{zhang2024navid, zhou2024navgpt}, and autonomous driving~\cite{guo2024vlm, chen2024driving}. 
A shared requirement across these applications is 
\textit{spatial intelligence}, i.e., understanding and reasoning
about the 3D structure of the physical world.
Despite excelling at textual and 2D visual understanding tasks, 
existing MLLMs remain limited in 3D spatial 
reasoning~\cite{yang2025thinking, chen2024spatialvlm}.

\begin{figure*}[t!]
  \begin{center}
    \centerline{\includegraphics[width=0.95\columnwidth]{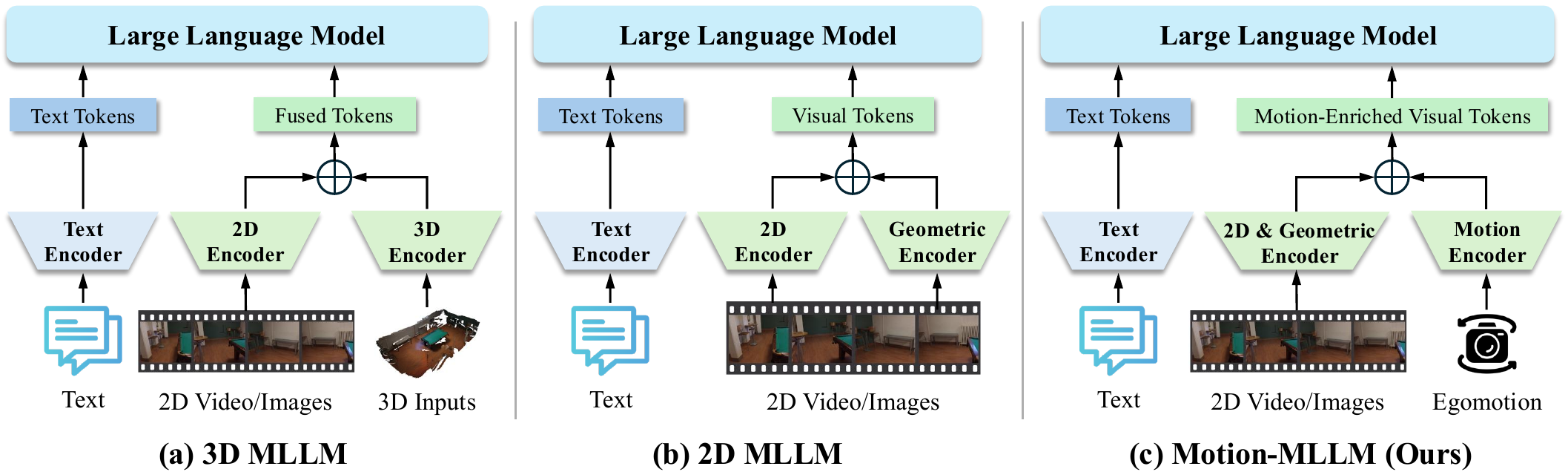}}
    \vspace*{-0.1in}
    \caption{
      Comparison of (a) 3D-input, (b) 2D-input, and (c) our 
      egomotion-input approaches for spatial reasoning in MLLMs.
    }
    \label{fig:intro}
  \end{center}
  \vspace*{-0.4in}
\end{figure*}

Existing studies on 3D spatial reasoning in MLLMs generally 
follow two directions, as shown in Fig.~\ref{fig:intro}. 
One line of work incorporates explicit 3D data, such as point 
clouds~\cite{hong20233d, xu2024pointllm, chen2024ll3da}, depth 
maps~\cite{zhu2024llava, zheng2025video}, or BEV 
maps~\cite{qi2025gpt4scene}, into the model. These methods 
achieve effective spatial awareness but rely on specialized 
sensors or computation-intensive processing pipelines. Another line of work avoids explicit 3D 
input and instead extracts geometric features directly from 2D 
video~\cite{ouyang2025spacer, li2025videochat, zheng2025learning, wu2025spatial}. 
While more practical, they cannot recover reliable metric scale 
from monocular geometry, thus limiting their capability 
to resolve ambiguities in distance and size.

In this paper, we introduce egomotion as a new input modality 
for MLLMs (Fig.~\ref{fig:intro}), captured by low-cost Inertial 
Measurement Units (IMUs) that are widely deployed on video-capturing 
platforms such as smartphones, 
robots, and vehicles~\cite{brossard2020ai}. Our motivation stems 
from how humans perceive their surroundings: 
we do not rely on vision only but continuously integrate 
\textit{bodily motion cues} to understand the space. 
For instance, eye movements and head rotation provide information
on the relative positions of objects we see, while the distance 
we travel helps us judge the scale of a room or the size of an 
object. Similarly, the egomotion data recorded by IMUs measures 
physical movements, providing reliable \blue{metric cues} that ground
visual observations in the real-world scale. This allows MLLMs 
to resolve the distance and size ambiguities that limit 
vision-only approaches, without incurring the overhead of 
explicit 3D representations.

We propose \sysname, a novel framework built on two key components 
designed to synergistically leverage the motion modality. 
First, we introduce a \textit{cascaded motion-visual keyframe 
filtering module} that selects a sparse yet representative set of 
keyframes from the video sequence. It employs a three-stage 
cascade progressing from lightweight egomotion checks to demanding 
visual feature analysis, so that computationally expensive 
operations are performed only for a small subset of informative 
frames. We then design an \textit{asymmetric cross-modal fusion 
module} that integrates visual and egomotion features through 
two-layer cross-attention fusion. Specifically, we adopt a GRU 
(Gated Recurrent Unit) based encoder to compress variable-length 
IMU segments between keyframes into motion tokens, 
which are then fused with 
visual tokens through a bidirectional attention layer followed by 
a unidirectional layer. In this design, motion tokens serve as 
intermediaries that channel egomotion cues and cross-frame 
context into the visual representation, producing 
egomotion-enriched visual tokens. By grounding visual content 
in physical egomotion trajectories, \sysname enables LLMs to reason about absolute scale and spatial 
relationships across the scene.

Extensive experimentation on five 3D scene understanding and spatial reasoning benchmarks shows that \sysname achieves an overall score of \blue{58.2} on VSI-Bench~\cite{yang2025thinking}, a \blue{$+7.5$} improvement over the SOTA despite using only $\sim$4B parameters.
On ScanQA~\cite{azuma2022scanqa} and SQA3D~\cite{ma2022sqa3d}
benchmarks, \sysname significantly outperforms all 2D-input
baselines and is competitive with SOTA 3D-input methods. 
On visual grounding (ScanRefer
\cite{chen2020scanrefer}) and dense captioning (Scan2Cap
\cite{chen2021scan2cap}) tasks, \sysname rivals or even 
outperforms 3D-input methods without requiring explicit 3D data. 
\blue{Moreover, \sysname runs $1.30\times$ and $1.61\times$ faster
than SOTA 2D- and 3D-input methods at matched accuracy, respectively, 
demonstrating that egomotion grounding delivers stronger spatial 
reasoning than vision-only approaches while avoiding the overhead 
of explicit 3D representations.}




\section{Related Work}
\vspace*{-0.05in}

\noindent\textbf{3D MLLMs for Scene Understanding.}
Recent advances extend LLMs for 3D scene 
understanding by incorporating explicit 3D representations 
such as point clouds, depth maps, or reconstructed 3D 
structures~\cite{hong20233d, xu2024pointllm, guo2023point, 
chen2024ll3da, wang2023chat, huang2024chat, huang2023embodied, 
huang2025leo, zhu2024llava, qi2025gpt4scene, zheng2025video}. 
PointLLM~\cite{xu2024pointllm} and Point-Bind
\cite{guo2023point} use point cloud encoders to let LLMs 
perceive 3D objects. LL3DA~\cite{chen2024ll3da} operates on 
point cloud scenes and further supports interactive visual 
prompts such as 3D bounding boxes. Chat3D~\cite{wang2023chat} 
and Chat-Scene~\cite{huang2024chat} adopt object-level scene 
representations to bridge 3D scenes with language.
LLaVA-3D~\cite{zhu2024llava} projects 2D features into 
3D space using depth maps, 
GPT4Scene~\cite{qi2025gpt4scene} constructs BEV maps through 3D 
reconstruction, and Video-3D LLM~\cite{zheng2025video} injects 3D 
position encodings into video representations. While effective, 
these methods rely on point clouds, depth maps, or 3D 
reconstruction inputs, which are expensive to acquire or 
computationally heavy. In contrast, our method uses lightweight 
egomotion data from IMU sensors to provide metric spatial grounding.

\noindent\textbf{2D MLLMs for Scene Understanding.}
Another line of work enhances spatial reasoning using only 2D images 
or videos, avoiding explicit 3D input~\cite{ouyang2025spacer, li2025videochat, zheng2025learning, wu2025spatial}. 
SpaceR~\cite{ouyang2025spacer} and VideoChat~\cite{li2025videochat} 
attempt to elicit 3D reasoning directly from existing VLMs without 
introducing additional geometry encoders. VG-LLM~\cite{zheng2025learning} 
employs a visual geometry encoder to extract 3D prior information from 
video sequences. Spatial-MLLM~\cite{wu2025spatial} extracts geometric 
features from video to boost spatial reasoning. These approaches reduce 
the dependency on 3D inputs but remain limited in resolving ambiguities 
in absolute scale and distance. Our method addresses this limitation by 
augmenting geometric features with egomotion data from IMU sensors, 
which provides the absolute metric scale that monocular geometry 
cannot reliably recover.

\noindent\textbf{Motion Modality for Scene Understanding.}
Motion has long been a key part of 3D vision, especially in areas like 
autonomous driving and robotics~\cite{campos2021orb, keetha2024splatam, yan2024gs, gholami2025spatial, xu2025egodtm}. Traditional systems
like ORB-SLAM3~\cite{campos2021orb} and newer neural approaches like 
SplaTAM~\cite{keetha2024splatam} or GS-SLAM~\cite{yan2024gs} show how 
tracking camera movement is essential for building high-fidelity maps. In the MLLM field, Omni-modality models like 
PandaGPT~\cite{su2023pandagpt} and One-LLM~\cite{han2024onellm} 
have begun incorporating IMU data to expand their reasoning. 
SensorLLM~\cite{li2025sensorllm} and LLM4HAR~\cite{hong2025llm4har} 
align IMU data with text, though they mostly focus on recognizing 
human activities. Newer frameworks like HIS-GPT~\cite{zhao2025his} 
and VMRMOT~\cite{lv2025vision} have also integrated motion 
dynamics to better capture human-scene interactions and improve 
multi-object tracking. In contrast, our work treats IMU data as 
a critical signal for both visual filtering and physical grounding. 
By integrating egomotion
data into the MLLM reasoning process, we provide \blue{lightweight 
yet reliable metric cues} that vision-only models lack. 
\blue{To the best of our knowledge, we are the first to use 
concurrent camera egomotion for scene-level spatial reasoning in MLLMs.}

\section{Design of \sysname}
\label{sec:design}
\vspace*{-0.05in}

\begin{figure*}[t!]
  \begin{center}
    \centerline{\includegraphics[width=0.95\columnwidth]{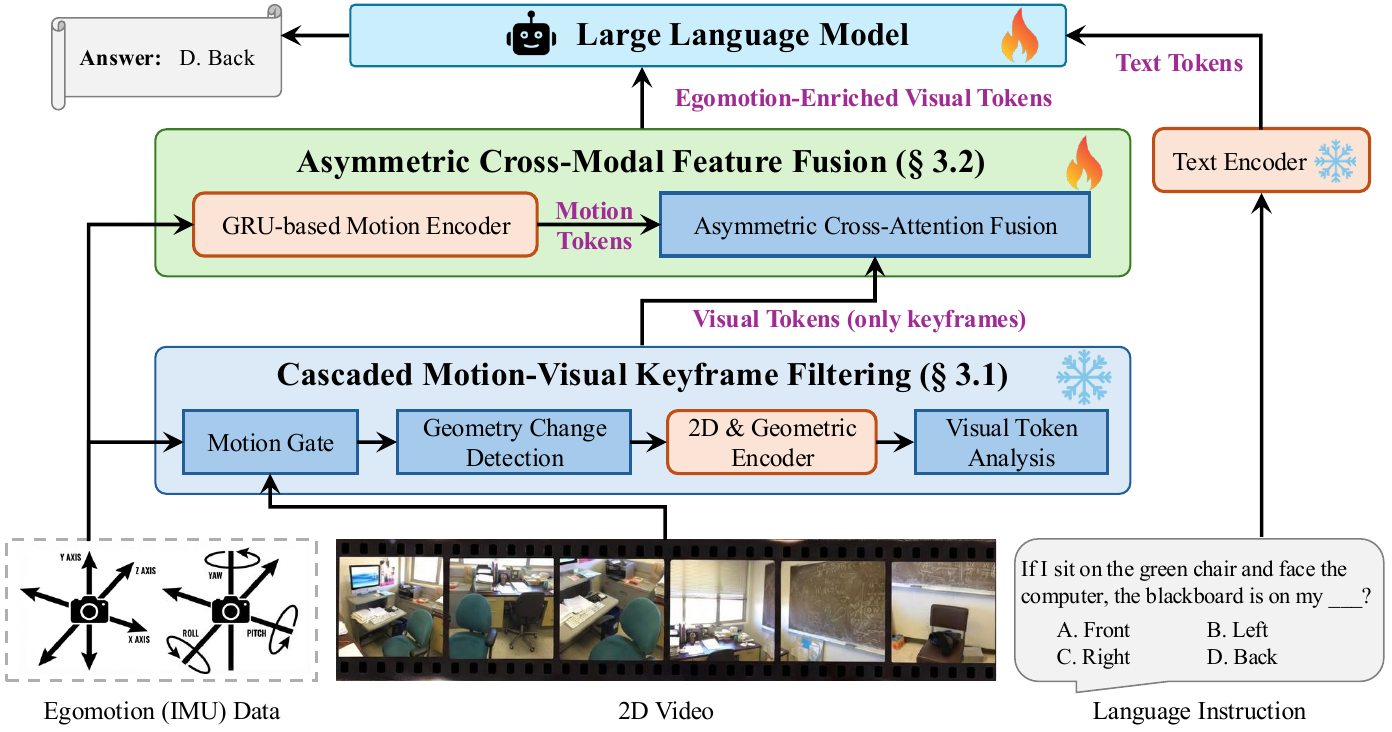}}
    \vspace*{-0.1in}
    \caption{
      \textbf{Architecture of \sysname.} 
      The \textcolor{orange}{\faFire} and 
      \textcolor{cyan}{\faSnowflake} icons indicate trainable 
      and frozen modules, respectively.
    }
    \label{fig:architecture}
  \end{center}
  \vspace*{-0.4in}
\end{figure*}

We propose a framework, \sysname, that enhances MLLMs for
3D scene understanding by introducing egomotion as an explicit 
input modality alongside visual observations.
Fig.~\ref{fig:architecture} shows the overall architecture 
of \sysname.
It takes two synchronized inputs: 2D video streams and 
concurrently captured IMU data that records camera motion. 
IMU sensors are commonly available on widely deployed 
video-capturing platforms, including smartphones, robotic 
systems, and autonomous vehicles. We detail two key 
components of \sysname: (1) a cascaded motion-visual keyframe 
filtering module (Sec.~\ref{subsec:keyframe-filtering}) that 
selects information-rich frames from the video stream in a 
computationally efficient manner; and (2) an asymmetric 
cross-modal feature fusion module (Sec.~\ref{subsec:cross-modal}) 
that integrates visual and motion features through two-layer 
cross-modal attention.

\subsection{Cascaded Motion-Visual Keyframe Filtering}
\label{subsec:keyframe-filtering}
\vspace*{-0.05in}

Due to the limited GPU memory and the high redundancy in 
consecutive video frames, MLLMs can typically process only 
a small subset of frames from a video. A common 
solution~\cite{bai2025qwen25vltechnicalreport,yang2025thinking} 
is uniform sampling, which often wastes the frame budget on static or repetitive segments and misses brief, dynamic events.
Some recent works~\cite{zheng2025video,wu2025spatial} propose 
to select frames whose geometric features maximally cover the 
3D scene. These methods extract 3D features from 
a large number of candidate frames and rely on additional 
3D inputs such as depth maps, incurring 
substantial overhead.

To overcome these limitations, we introduce a motion-visual 
keyframe filtering module, which selects frames based on both 
camera motion and visual content. Our key insight is that egomotion data provides a lightweight criterion for keyframe selection: a keyframe should be selected when it corresponds to a significant change in camera pose.
To implement this efficiently, we propose a cascaded filtering 
pipeline. Instead of evaluating all criteria on every frame, this pipeline progresses from inexpensive checks to demanding analyses, reserving expensive computations (e.g., visual feature extraction) for a small subset of candidate frames. The filtering proceeds in up to three stages per frame (see Sec.~\ref{appendix:keyframe-filtering} for the algorithm details and \blue{threshold settings}):

\noindent\textbf{Stage 1: Motion Gate.}
The IMU sensor records 6-axis inertial data (3-axis accelerometer 
and 3-axis gyroscope) concurrently with the video, providing 
measurements between consecutive frames. 
For a candidate frame $f_t$, the translational displacement
$d(\hat{f}_j, f_t)$ and the rotation angle $\theta(\hat{f}_j, f_t)$ 
since the most recently selected keyframe $\hat{f}_j$ 
can be easily obtained by integrating 
the accelerometer and gyroscope readings. 
A frame $f_t$ is discarded if $d(\hat{f}_j, f_t) < \tau_d$ 
and $\theta(\hat{f}_j, f_t) < \tau_\theta$, where $\tau_d$ 
and $\tau_\theta$ are predefined translation and rotation 
thresholds. This check is very simple and fast but can 
discard the vast majority of redundant frames where the 
camera is static or moves very slowly.

\noindent\textbf{Stage 2: Lightweight Geometric Change Detection.}
This stage performs a lightweight check on the frames that pass
Stage~1. It employs a sparse feature tracker from a SLAM 
front end~\cite{campos2021orb} to avoid full feature extraction. 
Specifically, it calculates the average parallax of features that 
are tracked in the last keyframe. A high parallax value indicates 
a substantial geometric change in the frame content.
If the parallax is not significant, the frame is discarded. Otherwise, it proceeds to Stage~3.

\noindent\textbf{Stage 3: Visual Token Analysis.}
This stage first extracts visual features from each candidate 
frame $f_t$ that passed the preceding checks. 
Following~\cite{wu2025spatial}, we obtain a fused visual 
token $\mathbf{v}_t$ by integrating 2D features 
from the MLLM's 2D visual encoder (e.g., Qwen2.5-VL's visual 
encoder~\cite{bai2025qwen25vltechnicalreport}) with 
geometric features from the VGGT backbone~\cite{wang2025vggt}. 
Since the cascaded pipeline filters out most frames in the 
first two stages, both 2D and geometric encoders run only on
the small subset that passed the inexpensive egomotion
and parallax checks. We then compute the cosine distance between $\mathbf{v}_t$ and the last keyframe's token $\mathbf{v}_j$.
If the distance exceeds a threshold $\tau_v$, the candidate 
is selected as a new keyframe. In particular, the 
$N$ final selected keyframes yield visual tokens $\mathbf{V}
= \left\{\mathbf{v}_1, \ldots, \mathbf{v}_N\right\}$ that 
are directly input to the cross-modal fusion module 
(Sec.~\ref{subsec:cross-modal}), 
avoiding redundant feature extraction.

\vspace*{-0.05in}
\subsection{Asymmetric Cross-Modal Feature Fusion}
\label{subsec:cross-modal}
\vspace*{-0.05in}

\begin{figure}[t!]
  \begin{center}
    \centerline{\includegraphics[width=0.9\columnwidth]{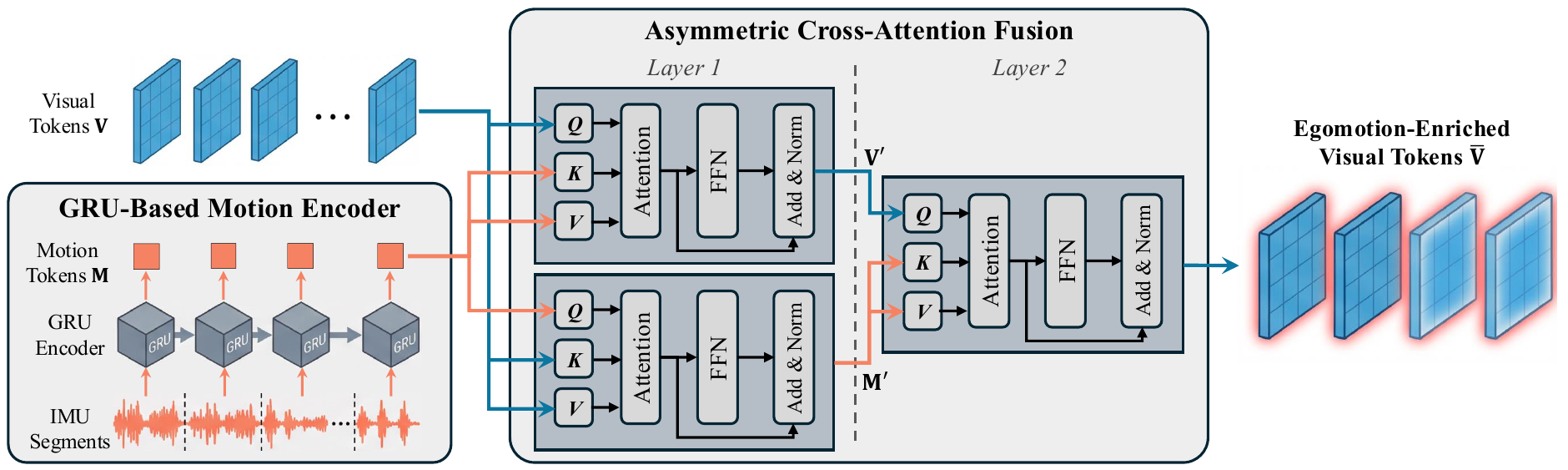}}
    \vspace*{-0.05in}
    \caption{
      Illustration of asymmetric cross-modal feature fusion.
    }
    \label{fig:fusion}
  \end{center}
  \vspace*{-0.4in}
\end{figure}

While the keyframe filtering module reduces redundancy in 
the input video, we need to integrate visual and egomotion 
information into an egomotion-aware video representation for 
3D scene understanding. 
As illustrated in Fig.~\ref{fig:fusion}, we first encode 
IMU data segments into compact motion features. 
We then design an asymmetric cross-attention mechanism that 
fuses them with the visual features.

\vspace*{-0.05in}
\subsubsection{GRU-based Motion Encoder.}
\vspace*{-0.05in}
Between consecutive keyframes $\hat{f}_{i-1}$ and $\hat{f}_i$, 
the IMU segment $S_i \in \mathbb{R}^{L_i \times 6}$ has variable 
length $L_i$ because keyframes are non-uniformly selected.
\blue{We employ a 2-layer bidirectional Gated Recurrent Unit 
(GRU)~\cite{cho2014learning} with hidden size 256 to encode each segment. 
GRU-based encoders naturally handle variable-length temporal sequences 
and remain the standard choice for IMU sequence encoding in inertial 
navigation~\cite{chen2018ionet, herath2020ronin}. We take the final
hidden state as the motion token $\mathbf{m}_i$, which naturally 
summarizes the cumulative egomotion integrated over the IMU segment 
between consecutive keyframes}.
For the first keyframe $\hat{f}_1$, we introduce a learnable start token 
$\mathbf{m}_1$. The resulting motion tokens form $\mathbf{M} = 
\left\{\mathbf{m}_1, \ldots, \mathbf{m}_N\right\}$, where $N$ is 
the number of selected keyframes. Each motion token has 
a one-to-one correspondence with the visual tokens from the
keyframe filtering stage.

\vspace*{-0.05in}
\subsubsection{Asymmetric Two-Layer Cross-Attention Fusion.}
\vspace*{-0.05in}

We design a two-layer cross-attention fusion module to integrate the 
visual and egomotion modalities. As shown in Fig.~\ref{fig:fusion}, this 
module takes visual tokens $\mathbf{V}$ and motion tokens $\mathbf{M}$
and produces egomotion-enriched visual tokens $\mathbf{\bar{V}}$, which 
maintains the same dimensionality as $\mathbf{V}$ to minimize the 
impact on the MLLM's feature space.

We first apply a bidirectional cross-attention layer where both 
modalities query each other. In the motion-to-vision direction, motion 
tokens serve as keys and values, providing absolute metric scale and 
egomotion trajectories that complement the geometric information in the 
visual tokens. In the vision-to-motion direction, visual tokens serve as 
keys and values to enrich motion features with visual context about the 
scenes. The two attention operations are computed as:
\begin{equation}
\begin{aligned}
\mathbf{V}' &= \mathbf{V} + \mathrm{FFN}(\mathrm{Attn}(\mathbf{V}W_Q^{v},\; \mathbf{M}W_K^{m},\; \mathbf{M}W_V^{m})) \\
\mathbf{M}' &= \mathbf{M} + \mathrm{FFN}(\mathrm{Attn}(\mathbf{M}W_Q^{m},\; \mathbf{V}W_K^{v},\; \mathbf{V}W_V^{v}))
\end{aligned}
\end{equation}
where $\mathrm{Attn}(Q,K,V) = \mathrm{softmax}(\frac{QK^T}{\sqrt{d_k}})V$, all $W$ are learnable projection matrices, and $\mathrm{FFN}(\cdot)$ denotes a feed-forward network~\cite{vaswani2017attention} applied with a residual connection. \blue{Both attention layers apply rotary positional embeddings (RoPE)~\cite{su2024roformer} to queries and keys, with $\mathbf{v}_i$ and $\mathbf{m}_i$ sharing position index $i$ to align each motion token with its paired keyframe and preserve cross-frame temporal order.}

Next, we fuse $\mathbf{V}'$ and $\mathbf{M}'$ with a unidirectional 
cross-attention layer. Only visual tokens query motion tokens in 
this layer, creating a visual$\rightarrow$motion$\rightarrow$visual 
information pathway. 
Since $\mathbf{M}'$ already carries visual context from other keyframes 
(absorbed in the first layer), this unidirectional attention layer 
enables motion-guided inter-frame visual communication, where motion 
tokens act as bridges that relay visual information across frames along 
physically grounded egomotion trajectories. The attention is computed as:
\begin{equation}
\mathbf{\bar{V}} = \mathbf{V}' + \mathrm{FFN}(\mathrm{Attn}(\mathbf{V}'W_Q,\; \mathbf{M}'W_K,\; \mathbf{M}'W_V))
\end{equation}
This asymmetric design retains only visual tokens as the output, 
while motion tokens are discarded after channeling 
cross-frame information into the visual tokens. Finally, the enriched 
visual tokens $\mathbf{\bar{V}}$ are fed into the MLLM alongside 
the text prompt through the standard visual-language pipeline.

\section{Experimental Evaluation}
\label{sec:experiments}
\vspace*{-0.05in}

\subsection{Experimental Setup}
\label{subsec:eval_setup}
\vspace*{-0.05in}

\noindent\textbf{Datasets.} 
For spatial reasoning, our model is tested on ScanQA~\cite{azuma2022scanqa}, SQA3D~\cite{ma2022sqa3d}, and VSI-Bench~\cite{yang2025thinking}. For the visual grounding task, our model is tested on ScanRefer~\cite{chen2020scanrefer}, which aims to locate the target object's bounding box in camera coordinates along with its corresponding frame index. For the dense captioning task, we utilize the Scan2Cap~\cite{chen2021scan2cap} benchmark, which requires generating descriptive captions for all objects within a scene. \sysname takes egomotion data as input, which is absent from these existing datasets. \blue{We therefore synthesize camera egomotion data in the format of a typical IMU sensor output, following standard practice when such sensor data is unavailable. We use a unified synthesis pipeline across all five benchmarks: cubic B-splines~\cite{lovegrove2013spline} fit to source-dataset ground-truth camera poses are analytically differentiated and sampled at 200\,Hz, then augmented with a standard MEMS noise model~\cite{forster2016manifold} to match real IMU sensors in both format and noise statistics. For VSI-Bench, the official per-video metadata~\cite{yang2025thinking} links each clip to its source scan in ScanNet, ScanNet++, or ARKitScenes, from which we obtain the ground-truth camera poses. These validation splits do not overlap with our training data. \sysname consumes only the 2D video frames and the noisy IMU stream at inference. Details of our egomotion data synthesis process are provided in Sec.~\ref{appendix:egomotion-data}. We further validate \sysname on a small-scale benchmark we construct from 6 TUM-VI~\cite{schubert2018tum} indoor sequences with real IMU recordings.}

\noindent\textbf{Implementation and training details.}
\sysname is built upon Qwen2.5-VL-3B~\cite{bai2023qwen} (4.3B parameters total). We leverage the 2D
visual encoder from Qwen2.5-VL~\cite{bai2025qwen25vltechnicalreport} 
and the VGGT backbone~\cite{wang2025vggt} as the geometric encoder 
to produce visual tokens $\mathbf{V}$, and Qwen2.5-VL's LLM backbone for text processing.
We train \sysname as a single generalist model on a mixture of
datasets (Sec.~\ref{appendix:training-data}), including the training
split of ScanQA~\cite{azuma2022scanqa}, SQA3D~\cite{ma2022sqa3d}, 
ScanRefer~\cite{chen2020scanrefer}, and Scan2Cap~\cite{chen2021scan2cap}. 
Each batch is randomly sampled from a single task type. For general tasks like question
answering, we use a standard cross-entropy loss. For the visual
grounding task, \blue{we use the same pre-extracted object proposals as VG~LLM~\cite{zheng2025learning} for fair comparison.} We follow~\cite{zheng2025video} to employ a specifically designed visual grounding loss~\cite{oord2018representation} to supervise proposal selection more accurately.
During training, we adopt a two-stage strategy. In the first stage, we freeze the 2D visual encoder, the VGGT backbone, and the LLM backbone, training only our newly introduced motion encoder and cross-attention modules using a learning rate of 1e-4 for one epoch. This allows the model to first learn to encode egomotion data and fuse it with visual features. In the second stage, we continue to keep the visual encoder and the VGGT backbone frozen but unfreeze the LLM backbone, fine-tuning the motion encoder, cross-attention modules, and the LLM end-to-end for another epoch. For this main training stage, we utilize the Adam optimizer~\cite{kingma2014adam} with a global batch size of 32. We warm up the learning rate to a peak of 1e-5 over the first 3\% of steps, then decay linearly. We preprocess the video for each scan to $640\times480$ resolution. 
Training uses 8 NVIDIA RTX 4000 Ada GPUs and takes $\sim$15 hours. All result numbers are averaged over three independent runs with different random seeds.


\vspace*{-0.05in}
\subsection{Spatial Reasoning Benchmarks}
\label{subsec:eval_spatial_reasoning}
\vspace*{-0.05in}

\begin{table}[t!]
\centering
\caption{\textbf{Evaluation Results on ScanQA~\cite{azuma2022scanqa} and SQA3D~\cite{ma2022sqa3d}.} ``2D'', ``3D'', and ``M'' specify the model's input type as 2D data, 3D data, and egomotion data, respectively.}
\label{tab:ScanQA}
\resizebox{.9\textwidth}{!}{%
\begin{tabular}{@{}l|ccc|ccccc|cc@{}}
\toprule
\multirow{2}{*}{\textbf{Methods}} & \multicolumn{3}{c|}{\textbf{Input}} & \multicolumn{5}{c|}{\textbf{ScanQA (val)}}                                     & \multicolumn{2}{c}{\textbf{SQA3D (test)}} \\ \cmidrule(l){2-11}
                                  & 2D          & 3D        & M         & BLEU-1        & BLEU-4        & METEOR        & ROUGE-L       & CIDEr          & EM@1                 & EM@R1               \\ \midrule
\textit{Task-Specific Models}     &             &           &           &               &               &               &               &                &                      &                     \\
SQA3D~\cite{ma2022sqa3d}                             &            & \Checkmark &           & \textbf{30.5} & \textbf{11.2} & 13.5          & 34.5          & -              & 46.6                 & -                   \\
3D-VisTA~\cite{zhu20233d}                          &            & \Checkmark &           & -             & 10.4          & \textbf{13.9} & \textbf{35.7} & \textbf{69.6}  & \textbf{48.5}        & -                   \\ \midrule
\textit{2D-Input Models}       &             &           &           &               &               &               &               &                &                      &                     \\
Qwen2.5-VL-3B~\cite{bai2025qwen25vltechnicalreport}                     & \Checkmark  &           &           & 22.5          & 3.8           & 9.7           & 25.4          & 47.4           & 43.4                 & 45.9                \\
Qwen2.5-VL-7B~\cite{bai2025qwen25vltechnicalreport}                      & \Checkmark  &           &           & 27.8          & 3.0           & 11.4          & 29.3          & 53.9           & 46.5                 & 49.8                \\
Spatial-MLLM~\cite{wu2025spatial}                      & \Checkmark  &           &           & \textbf{44.4} & \textbf{14.8} & \textbf{18.4} & \textbf{45.0} & \textbf{91.8}  & \textbf{55.9}        & \textbf{58.7}       \\  \midrule
\textit{3D/2.5D-Input Models}     &             &           &           &               &               &               &               &                &                      &                     \\
LL3DA~\cite{chen2024ll3da}                             &            & \Checkmark &           & -             & 13.5          & 15.9          & 37.3          & 76.8           & -                    & -                   \\
GPT4Scene-HDM~\cite{qi2025gpt4scene}                               & \Checkmark  & \Checkmark &           & 44.4             & 15.5          & 18.9          & 46.5          & 96.3           & \textbf{60.6}                 & \textbf{63.3}                   \\
Video-3D LLM~\cite{zheng2025video}                      & \Checkmark  & \Checkmark &           & \textbf{47.1} & 16.2          & 19.8          & 49.0          & 102.1          & 58.6                 & -                   \\
LLaVA-3D~\cite{zhu2024llava}                          & \Checkmark  & \Checkmark &           & -             & \textbf{16.4} & \textbf{20.8} & \textbf{49.6} & \textbf{103.1} & 60.1        & -                   \\ \midrule
\textit{Egomotion-Input Model}    &             &           &           &               &               &               &               &                &                      &                     \\
\textbf{\sysname-4B}                       & \Checkmark  &           & \Checkmark & \textbf{46.5} & \textbf{15.8} & \textbf{20.3} & \textbf{48.0} & \textbf{100.4} & \textbf{60.2}        & \textbf{62.4}       \\ \bottomrule
\end{tabular}%
}
\vspace*{-0.2in}
\end{table}

\noindent\textbf{Evaluation on ScanQA and SQA3D.}
ScanQA~\cite{azuma2022scanqa} and SQA3D~\cite{ma2022sqa3d} are two 3D question-answering benchmarks built upon ScanNet~\cite{dai2017scannet} indoor scenes. We evaluate \sysname on the ScanQA validation set (4,675 QA pairs) and on the SQA3D test set (3,519 QA pairs). As shown in Tab.~\ref{tab:ScanQA}, \sysname outperforms task-specific and 2D-input models across all metrics, achieving $+8.6$ CIDEr and $+4.3$ EM@1 over the SOTA 2D-input baseline (Spatial-MLLM). This shows that the egomotion modality enhances spatial and positional reasoning. Among 3D/2.5D-input models, only Video-3D LLM~\cite{zheng2025video} (on ScanQA), GPT4Scene~\cite{qi2025gpt4scene} (on SQA3D), and LLaVA-3D~\cite{zhu2024llava} (on ScanQA) surpass \sysname, but these models rely on point clouds or depth maps that incur high data collection and computation costs. \sysname achieves comparable performance to the SOTA 3D-input models LLaVA-3D and GPT4Scene ($-2.7$ CIDEr and $-0.4$ EM@1) using low-cost and lightweight egomotion data. Complete ScanQA metrics and SQA3D results are provided in Sec.~\ref{appendix:scanqa-sqa3d}.

\begin{table}[t!]
\caption{\textbf{Evaluation Results on VSI-Bench~\cite{yang2025thinking}.} We follow the standard setting of VSI-Bench and utilize the keyframe filtering module to select frames for baselines and \sysname, respectively. \textbf{Bold} and \underline{underline} denote the best and second-best results in each column, respectively.}
\centering
\label{tab:VSI-Bench}
\resizebox{\textwidth}{!}{%
\begin{tabular}{@{}l|cccccccc|c@{}}
\toprule
\multirow{2}{*}{\textbf{Methods}} & \multicolumn{4}{c}{\textbf{Numerical Question}}                                    & \multicolumn{4}{c|}{\textbf{Multiple-Choice Question}}        & \multirow{2}{*}{\textbf{Avg.}} \\ \cmidrule(lr){2-9}
                                  & Obj. Cnt.     & Abs. Dist.    & Obj. Size     & \multicolumn{1}{c|}{Room Size}     & Rel. Dist.    & Rel. Dir.     & Route Plan    & Appr. Order.  &                                \\ \midrule
\textit{Proprietary Models (API)} &               &               &               & \multicolumn{1}{c|}{}              &               &               &               &               &                                \\
GPT-4o~\cite{achiam2023gpt}                            & 46.2          & 5.3           & 43.8          & \multicolumn{1}{c|}{38.2}          & 37.0          & 41.3          & 31.5          & 28.5          & 34.0                           \\
Gemini-1.5 Pro~\cite{team2024gemini}                    & 56.2          & 30.9          & \underline{64.1} & \multicolumn{1}{c|}{43.6}       & 51.3          & \underline{46.3} & \textbf{36.0} & 34.6          & 45.4                           \\ \midrule
\textit{Open-source Models}       &               &               &               & \multicolumn{1}{c|}{}              &               &               &               &               &                                \\
LLaVA-OneVision-72B~\cite{li2024llava}               & 43.5          & 23.9          & 57.6          & \multicolumn{1}{c|}{37.5}          & 42.5          & 39.9          & 32.5          & 44.6          & 40.2                           \\
Qwen2.5-VL-3B~\cite{bai2025qwen25vltechnicalreport}                      & 24.3          & 24.7          & 31.7          & \multicolumn{1}{c|}{22.6}          & 38.3          & 41.6          & 26.3          & 21.2          & 30.6                           \\
Qwen2.5-VL-7B~\cite{bai2025qwen25vltechnicalreport}                      & 40.9          & 14.8          & 43.4          & \multicolumn{1}{c|}{10.7}          & 38.6          & 38.5          & 33.0          & 29.8          & 33.0                           \\
InternVL3-78B~\cite{zhu2025internvl3}                     & \underline{71.2} & \textbf{53.7} & 44.4       & \multicolumn{1}{c|}{39.5}          & \underline{55.9} & 39.5       & 28.9          & 54.5          & 48.5                           \\ \midrule
\textit{Spatial Reasoning Models} &               &               &               & \multicolumn{1}{c|}{}              &               &               &               &               &                                \\
Spacer~\cite{ouyang2025spacer}                            & 57.8          & 28.2          & 59.9          & \multicolumn{1}{c|}{47.1}          & 40.1          & 45.4          & 33.5          & 52.1          & 45.5                           \\
VG LLM-4B~\cite{zheng2025learning}                         & 66.0          & 37.8          & 55.2          & \multicolumn{1}{c|}{59.2}          & 44.6          & 45.6          & 33.5          & 36.4          & 47.3                           \\
VG LLM-8B~\cite{zheng2025learning}                         & 67.9          & 37.7          & 58.6          & \multicolumn{1}{c|}{\underline{62.0}} & 46.6       & 40.7          & 32.4          & \underline{59.2} & \underline{50.7}            \\
Spatial-MLLM-4B~\cite{wu2025spatial}                   & 65.3          & 34.8          & 63.1          & \multicolumn{1}{c|}{45.1}          & 41.3          & 46.2          & 33.5          & 46.3          & 48.4                           \\ \midrule
\textbf{\sysname-4B}    & \textbf{72.2} & \underline{50.8} & \textbf{65.4} & \multicolumn{1}{c|}{\textbf{63.6}} & \textbf{62.2} & \textbf{55.4} & \underline{35.7} & \textbf{60.3} & \textbf{58.2}                  \\ \bottomrule
\end{tabular}%
}
\vspace*{-0.05in}
\end{table}

\noindent\textbf{Evaluation on VSI-Bench.}
VSI-Bench~\cite{yang2025thinking} contains over $5,000$
question-answer pairs curated from ScanNet~\cite{dai2017scannet}, 
ScanNet++~\cite{yeshwanth2023scannet++}, and 
ARKitScenes~\cite{baruch2021arkitscenes}, comprising eight subtasks.
As shown in Tab.~\ref{tab:VSI-Bench}, \sysname achieves the best score
on six of eight subtasks and the overall average, raising the average 
to \blue{58.2}, a \blue{$+7.5$} improvement over the SOTA baseline, 
outperforming proprietary and open-source models up to 78B with only 
$\sim$4B parameters. \blue{On the two exceptions (Abs.~Dist., Route Plan), 
\sysname trails by $\le 3$ points against models much larger than 
our $\sim$4B configuration}. \blue{We compare against Spatial-MLLM
\cite{wu2025spatial}, which also uses VGGT encoder but lacks our IMU 
encoder and cross-modal fusion. \sysname improves by $+16.0$ on the
cross-frame Abs.~Dist.\ task but only $+2.3$ on the intra-frame Obj.~Size 
task, attributing the gap to IMU-encoded inter-frame egomotion and 
asymmetric cross-modal fusion. It also benefits view-integration tasks 
(e.g., $+9.2$ on relative direction), where motion features bridge 
observations across viewpoints.}

\begin{table}[t!]
\caption{\blue{\textbf{Real IMU validation on TUM-VI~\cite{schubert2018tum}.} We follow the standard setting of VSI-Bench and utilize the keyframe filtering module to select frames for baselines and \sysname, respectively.}}
\centering
\label{tab:tumvi}
\resizebox{0.9\textwidth}{!}{%
\begin{tabular}{@{}l|cccccc|c@{}}
\toprule
\multirow{2}{*}{\textbf{Methods}} & \multicolumn{3}{c}{\textbf{Numerical Question}} & \multicolumn{3}{c|}{\textbf{Multiple-Choice Question}} & \multirow{2}{*}{\textbf{Avg.}} \\ \cmidrule(lr){2-7}
& Obj. Cnt. & Abs. Dist. & \multicolumn{1}{c|}{Room Size} & Rel. Dist. & Rel. Dir. & Appr. Order & \\ \midrule
Qwen2.5-VL-3B~\cite{bai2025qwen25vltechnicalreport} & 21.7 & 19.6 & \multicolumn{1}{c|}{20.8} & 35.0 & 35.0 & 20.0 & 25.4 \\
VG LLM-4B~\cite{zheng2025learning} & 50.4 & 25.4 & \multicolumn{1}{c|}{48.3} & 40.0 & 45.0 & 30.0 & 39.9 \\
Spatial-MLLM-4B~\cite{wu2025spatial} & 50.0 & 24.6 & \multicolumn{1}{c|}{38.3} & 40.0 & 45.0 & 35.0 & 38.8 \\ \midrule
\textbf{\sysname-4B} & \textbf{54.6} & \textbf{39.2} & \multicolumn{1}{c|}{\textbf{55.0}} & \textbf{45.0} & \textbf{55.0} & \textbf{45.0} & \textbf{49.0} \\ \bottomrule
\end{tabular}%
}
\vspace*{-0.05in}
\end{table}

\blue{\noindent\textbf{Real IMU Validation on TUM-VI.} We manually 
construct a small benchmark of 120 QA pairs over the 6 indoor sequences
of TUM-VI~\cite{schubert2018tum} (which provides real 200\,Hz BMI160 
IMU recordings), covering six tasks and following VSI-Bench
\cite{yang2025thinking} question templates (details are provided in 
Sec.~\ref{appendix:tumvi-details}). As shown in Tab.~\ref{tab:tumvi}, 
\sysname reaches an average of 49.0, $+9.1$ over the best-performing 
baseline (VG LLM-4B at 39.9), with the largest gains on Abs.\ Dist.\ 
($+13.8$) and Appr.\ Order ($+15.0$), confirming that \sysname's 
egomotion advantage transfers from synthesized to real IMU data.}

\vspace*{-0.05in}
\subsection{3D Scene Understanding Tasks} 
\label{subsec:eval_scene_understanding}
\vspace*{-0.05in}

We further assess \sysname's capabilities in visual grounding and 
dense captioning using the ScanRefer~\cite{chen2020scanrefer} and 
Scan2Cap~\cite{chen2021scan2cap} benchmarks, respectively. 
Unlike general question answering, these tasks differ fundamentally
by requiring explicit grounding mechanisms that link textual 
descriptions to specific 3D spatial structures. 
The quantitative results are summarized in Tab.~\ref{tab:ScanRefer}.

\begin{table}[t!]
\centering
\caption{\textbf{Evaluation of 3D scene understanding tasks.} For ScanRefer, scores include proposal refinement following SPAR~\cite{zhang2025flatland}; ``()'' indicates the score without refinement.}
\label{tab:ScanRefer}
\resizebox{0.9\textwidth}{!}{%
\begin{tabular}{@{}l|ccc|cc|cccc@{}}
\toprule
\multirow{2}{*}{\textbf{Methods}} & \multicolumn{3}{c|}{\textbf{Input}} & \multicolumn{2}{c|}{\textbf{ScanRefer}}     & \multicolumn{4}{c}{\textbf{Scan2Cap}}                        \\ \cmidrule(l){2-10}
                                  & 2D         & 3D        & M         & Acc@0.25$\uparrow$             & Acc@0.5$\uparrow$              & C@0.5$\uparrow$         & B-4@0.5$\uparrow$       & M@0.5$\uparrow$         & R@0.5$\uparrow$         \\ \midrule
\textit{Task-Specific Models}     &            &           &           &                      &                      &               &               &               &               \\
ScanRefer~\cite{chen2020scanrefer}                         &            & \Checkmark &           & 37.3                 & 24.3                 & -             & -             & -             & -             \\
Scan2Cap~\cite{chen2021scan2cap}                          &            & \Checkmark &           & -                    & -                    & 39.1          & 23.3          & 22.0          & 44.8          \\
3D-VisTA~\cite{zhu20233d}                          &            & \Checkmark &           & \textbf{50.6}        & \textbf{45.8}        & \textbf{66.9} & \textbf{34.0} &   -            &    -           \\ \midrule
\textit{2D-Input Models}          &            &           &           &                      &                      &               &               &               &               \\
SPAR-7B~\cite{zhang2025flatland}                           & \Checkmark  &           &           & 48.8 (31.9)          & 43.1 (12.4)          & -             & -             & -             & -             \\
VG LLM-4B~\cite{zheng2025learning}                         & \Checkmark  &           &           & 53.5 (36.4)          & 47.5 (11.8)          & 78.6          & 40.9          & 28.6          & 62.4          \\
VG LLM-8B~\cite{zheng2025learning}                         & \Checkmark  &           &           & \textbf{57.6 (41.6)} & \textbf{50.9 (14.9)} & \textbf{80.0} & \textbf{41.5} & \textbf{28.9} & \textbf{62.6} \\ \midrule
\textit{2.5D/3D-Input Models}     &            &           &           &                      &                      &               &               &               &               \\
Grounded 3D-LLM~\cite{chen2024grounded}                   &            & \Checkmark &           & 47.9                 & 44.1                 & 70.2          & 35.0          & -             & -             \\
GPT4Scene-HDM~\cite{qi2025gpt4scene}                     & \Checkmark  & \Checkmark &           & \textbf{62.6}        & \textbf{57.0}        & -             & 40.6          & -             & 59.3          \\
LLaVA-3D~\cite{zhu2024llava}                          & \Checkmark  & \Checkmark &           & 54.1                 & 42.4                 & 79.2          & \textbf{41.1} & \textbf{30.2} & \textbf{63.4} \\
Video-3D LLM~\cite{zheng2025video}                      & \Checkmark  & \Checkmark &           & 58.1                 & 51.7                 & \textbf{80.0} & 40.2          & 28.5          & 61.7          \\ \midrule
\textit{Egomotion-Input Models}   &            &           &           &                      &                      &               &               &               &               \\
\textbf{\sysname-4B}           & \Checkmark  &           & \Checkmark & \blue{\textbf{61.4 (45.8)}} & \blue{\textbf{55.3 (19.6)}} & \textbf{79.0} & \textbf{41.6} & \textbf{30.0} & \textbf{64.0} \\ \bottomrule
\end{tabular}%
}
\vspace*{-0.05in}
\end{table}


\noindent\textbf{Visual Grounding on ScanRefer.}
ScanRefer~\cite{chen2020scanrefer} contains $36,665$ object descriptions 
paired with axis-aligned bounding boxes across $562$ indoor scans. 
We follow the standard protocol with Acc@0.25 and Acc@0.5, the 
percentage of samples whose IoU exceeds the threshold.
As shown in Tab.~\ref{tab:ScanRefer}, \sysname achieves \blue{Acc@0.25/0.5 
of $61.4/55.3$ (refined) and $45.8/19.6$ (raw)} on the ScanRefer benchmark 
with only $\sim$4B parameters, without requiring external 3D data such as 
point clouds or depth maps. It significantly outperforms the SOTA 2D-input 
baseline (VG LLM-8B~\cite{zheng2025learning}), surpassing its refined 
scores by $+3.8$ and $+4.4$, respectively. More importantly, \sysname 
rivals and even outperforms methods relying on heavy explicit 3D inputs, 
such as LLaVA-3D ($54.1$ Acc@0.25) and Video-3D LLM ($58.1$ Acc@0.25).
This gain highlights the effectiveness of our asymmetric cross-modal 
fusion: motion features ground visual content in physical space, allowing 
\sysname to infer distances and sizes from egomotion cues without depth 
maps or point clouds.


\noindent\textbf{Dense Captioning on Scan2Cap.}
We evaluate \sysname on the val split of Scan2Cap~\cite{chen2021scan2cap}, 
which contains $9{,}508$ object descriptions across $141$ indoor scans. 
We report CIDEr (C), BLEU-4 (B-4), METEOR (M), and ROUGE (R) scores at 
IoU $\geq 0.5$ against the ground truth.
As shown in Tab.~\ref{tab:ScanRefer}, \sysname achieves a B-4@0.5 score
of $41.6$ and an R@0.5 score of $64.0$, setting a new best across all 
baselines, including those utilizing explicit 3D data. Its C@0.5 ($79.0$) 
and M@0.5 ($30.0$) are competitive, nearly matching top-performing
3D-input models such as Video-3D LLM~\cite{zheng2025video} and 
LLaVA-3D~\cite{zhu2024llava}. The improvement is smaller than on visual 
grounding, as dense captioning relies less on absolute scale estimation. 
However, gains on B-4, M, and R over 2D-input baselines demonstrate that 
leveraging the egomotion modality enables better object localization and 
spatial reasoning, yielding more precise and contextually grounded captions.

\vspace*{-0.05in}
\subsection{Cost-Effectiveness of \sysname}
\label{subsec:cost_effectiveness}
\vspace*{-0.05in}

\begin{table}[t!]
\centering
\caption{\textbf{Cost-effectiveness comparison on ScanQA~\cite{azuma2022scanqa} and SQA3D~\cite{ma2022sqa3d}.} ``MC'' represents the maximum coverage sampling~\cite{wu2025spatial,zheng2025video}. ``MV Filtering'' denotes the motion-visual keyframe filtering method we design. ``T'' represents the end-to-end time consumption.}
\label{tab:cost-effectiveness}
\resizebox{\textwidth}{!}{%
\begin{tabular}{@{}l|ccc|c|c|ccc|ccc@{}}
\toprule
\multirow{2}{*}{\textbf{Methods}} & \multicolumn{3}{c|}{\textbf{Input}}                                                  & \multirow{2}{*}{\textbf{\begin{tabular}[c]{@{}c@{}}Sampling \\ Strategy\end{tabular}}} & \multirow{2}{*}{\textbf{\# Frames}} & \multicolumn{3}{c|}{\textbf{ScanQA}}          & \multicolumn{3}{c}{\textbf{SQA3D}}            \\ \cmidrule(lr){2-4} \cmidrule(l){7-12} 
                                  & 2D                         & 3D                         & M                          &                                                                                        &                                     & EM(\%)$\uparrow$     & T($s$)$\downarrow$         & CE($s^{-1}$)$\uparrow$      & EM(\%)$\uparrow$     & T($s$)$\downarrow$         & CE($s^{-1}$)$\uparrow$      \\ \midrule
\multirow{2}{*}{Qwen2.5-VL-3B~\cite{bai2025qwen25vltechnicalreport}}    & \multirow{2}{*}{\Checkmark} & \multirow{2}{*}{}          & \multirow{2}{*}{}          & Uniform                                                                        & 128                                 & 17.1          & 2.45          & 0.07          & 49.4          & 2.39          & 0.21          \\
                                  &                            &                            &                            & Uniform                                                                       & 32                                  & 15.5          & \underline{0.72}    & 0.22          & 45.7          & \underline{0.71}    & 0.64    \\ \midrule
\multirow{3}{*}{Spatial-MLLM~\cite{wu2025spatial}}     & \multirow{3}{*}{\Checkmark} & \multirow{3}{*}{}          & \multirow{3}{*}{}          & Uniform                                                                        & 128                                 & 28.8          & 3.06          & 0.09          & \underline{62.2}    & 3.01          & 0.21          \\
                                  &                            &                            &                            & Uniform                                                                       & 32                                  & 26.2          & 0.88          & 0.30    & 58.5          & 0.87          & 0.67          \\
                                  &                            &                            &                            & MC                                                                       & $\sim$23                                  & 25.9          & 0.79          & \underline{0.33}    & 57.6          & 0.79          & \underline{0.73}          \\ \midrule
\multirow{3}{*}{Video-3D LLM~\cite{zheng2025video}}     & \multirow{3}{*}{\Checkmark} & \multirow{3}{*}{\Checkmark} & \multirow{3}{*}{}          & Uniform                                                                                & 128                                 & \underline{31.7}    & 3.45          & 0.09          & 60.8          & 3.36          & 0.18          \\
                                  &                            &                            &                            & Uniform                                                                                & 32                                  & 30.1          & 1.22          & 0.25          & 58.6          & 1.19          & 0.49          \\
                                  &                            &                            &                            & MC                                                                                     & $\sim$18                            & 29.5          & 0.98          & 0.30    & 57.7          & 0.97          & 0.59    \\ \midrule
\multirow{3}{*}{\textbf{\sysname-4B}}   & \multirow{3}{*}{\Checkmark} & \multirow{3}{*}{}          & \multirow{3}{*}{\Checkmark} & Uniform                                                                        & 128                                 & \textbf{31.5} & 3.10          & 0.10          & \textbf{62.8} & 3.05          & 0.21          \\
                                  &                            &                            &                            & Uniform                                                                       & 32                                  & 29.1          & 0.90          & 0.32          & 59.5          & 0.88          & 0.68          \\
                                  &                            &                            &                            & MV Filtering                                                                           & $\sim$21                            & 29.8          & \textbf{0.61} & \textbf{0.49} & 60.2          & \textbf{0.59} & \textbf{1.02} \\ \bottomrule
\end{tabular}%
}
\vspace*{-0.2in}
\end{table}

We evaluate the cost-effectiveness of \sysname on ScanQA
\cite{azuma2022scanqa} and SQA3D~\cite{ma2022sqa3d}.
Beyond the standard Exact Match (EM) accuracy and end-to-end inference
time $T$, \blue{we report speedup at matched accuracy against each 
baseline's best operating point, summarized by the ratio 
$\mathrm{CE}=\mathrm{EM}\%/T$ (higher is better). The full Pareto 
analysis is in Sec.~\ref{appendix:pareto}.} We compare \sysname against 
SOTA 2D- and 3D-input models under uniform sampling and method-specific 
adaptive strategies (Tab.~\ref{tab:cost-effectiveness}).

\noindent\textbf{Superior efficiency under uniform frame sampling.}
Under uniform sampling, more frames yield higher accuracy but incur 
significant computational overhead. However, \sysname demonstrates 
superior cost-effectiveness to both 2D- and 3D-input baselines.
Compared to 2D baselines (e.g., Spatial-MLLM~\cite{wu2025spatial}),
our model achieves significantly higher accuracy ($29.1\%$ vs.~$26.2\%$ 
on ScanQA with 32 frames) at comparable latency ($0.90$s vs.~$0.88$s).
\blue{Compared to 3D-input models such as Video-3D LLM
\cite{zheng2025video}, \sysname achieves comparable accuracy
($29.1$ vs.\ $30.1$ on ScanQA and $59.5$ vs.\ $58.6$ on SQA3D) while
running $\sim 1.36\times$ faster ($0.90$s vs.\ $1.22$s).} This efficiency 
stems from our lightweight motion encoder and cross-modal fusion, 
which avoid heavy 3D data processing.

\noindent\textbf{Benefits of egomotion-guided keyframe filtering.}
\blue{As shown in Tab.~\ref{tab:cost-effectiveness}, \sysname is both 
more accurate and faster than every baseline at its best operating point 
on both benchmarks. Against the SOTA 2D- and 3D-input baselines, 
\sysname runs $1.30\times$ and $1.61\times$ faster on ScanQA, respectively, 
with $+3.9$ and $+0.3$ EM accuracy gains, and similar margins on SQA3D. 
The CE metric results summarize this advantage as a single number.} 
Compared to 32-frame uniform sampling, our motion-visual keyframe filtering 
achieves comparable accuracy (e.g., $29.8\%$ vs.~$29.1\%$ on ScanQA) using 
only $\sim21$ keyframes, showing that egomotion effectively identifies 
information-rich frames. 

\vspace*{-0.05in}
\subsection{Ablation Study}
\label{subsec:ablation_study}
\vspace*{-0.05in}

\begin{table}[t!]
\centering
\caption{\textbf{Ablation study on ScanQA~\cite{azuma2022scanqa} and SQA3D~\cite{ma2022sqa3d}.} \blue{``VGGT-only'' corresponds to the Spatial-MLLM configuration. ``Visual-based Sampling'' filters frames purely based on visual criteria~\cite{wu2025spatial}.} ``Full MV Filtering'' refers to evaluating all criteria on every frame. ``Concat+MLP'' concatenates the visual and egomotion tokens and uses an MLP to transform them into the embedding space. \blue{All rows except the ``Qwen2.5-VL-3B'' are fine-tuned with the same two-stage pipeline as full \sysname.}}
\label{tab:ablation}
\resizebox{.75\textwidth}{!}{%
\begin{tabular}{@{}l|ccc|ccc@{}}
\toprule
\multicolumn{1}{c|}{\multirow{2}{*}{\textbf{Type}}} & \multicolumn{3}{c|}{\textbf{ScanQA}}          & \multicolumn{3}{c}{\textbf{SQA3D}}            \\ \cmidrule(l){2-7}
\multicolumn{1}{c|}{}                               & EM(\%)$\uparrow$     & T($s$)$\downarrow$         & CE($s^{-1}$)$\uparrow$      & EM(\%)$\uparrow$     & T($s$)$\downarrow$         & CE($s^{-1}$)$\uparrow$      \\ \midrule
Qwen2.5-VL-3B (Backbone)                            & 15.5          & 0.72          & 0.22          & 45.7          & 0.71          & 0.64          \\ \midrule
\blue{\textbf{\textit{Component Ablation}}}         &               &               &               &               &               &               \\
\blue{VGGT-only (no IMU)}                           & \blue{25.9}   & \blue{0.79}   & \blue{0.33}   & \blue{57.6}   & \blue{0.79}   & \blue{0.73}   \\
\blue{IMU-only (no VGGT)}                           & \blue{\underline{27.5}} & \blue{\underline{0.70}} & \blue{\underline{0.39}} & \blue{\underline{58.5}} & \blue{\underline{0.68}} & \blue{\underline{0.86}} \\ \midrule
\textbf{\textit{Frame Sampling Strategy}}           &               &               &               &               &               &               \\
Uniform 32-frame                                    & 29.1          & \underline{0.90} & \underline{0.32} & 59.5          & \underline{0.88} & \underline{0.68} \\
\blue{Visual-based Sampling (no IMU)}               & \blue{29.4}   & \blue{0.92}   & \blue{\underline{0.32}} & \blue{59.8}   & \blue{0.89}   & \blue{0.67}   \\
Full MV Filtering                                   & \underline{29.8} & 0.96          & 0.31          & \underline{60.3} & 0.93          & 0.65          \\ \midrule
\textbf{\textit{Feature Connector}}                 &               &               &               &               &               &               \\
Concat+MLP \blue{(95M)}                             & 19.2          & \underline{0.58} & 0.33          & 41.5          & \underline{0.55} & 0.75          \\
Single-Layer Cross-Attention \blue{(57M)}           & \underline{26.1} & 0.60          & \underline{0.44} & \underline{55.2} & 0.57          & \underline{0.96} \\ \midrule
\multicolumn{7}{@{}l}{\blue{\textbf{\textit{Ours: Cascaded MV Filtering + Asymmetric Cross-Attention (110M)}}}} \\
\blue{Full \sysname}                                & \blue{\textbf{29.8}} & \blue{\textbf{0.61}} & \blue{\textbf{0.49}} & \blue{\textbf{60.2}} & \blue{\textbf{0.59}} & \blue{\textbf{1.02}} \\ \bottomrule
\end{tabular}%
}
\vspace*{-0.15in}
\end{table}

\blue{\noindent\textbf{Effectiveness of egomotion modality.} 
Tab.~\ref{tab:ablation} disentangles the contributions of egomotion and 
VGGT-derived geometric features. Removing IMU (VGGT-only) drops accuracy 
by $-3.9\%/-2.6\%$ on ScanQA/SQA3D, while removing VGGT (IMU-only) drops 
by $-2.3\%/-1.7\%$. IMU-only outperforms VGGT-only on both benchmarks 
while also being faster, indicating that the metric inter-frame motion 
supplied by IMU drives the substantial performance gain.}

\noindent\textbf{Effectiveness of cascaded motion-visual keyframe 
filtering.} ``Full MV Filtering'' improves accuracy over uniform sampling 
to $29.8\%$ on ScanQA but increases latency to $0.96$s due to evaluating 
all criteria on every frame. \blue{Replacing IMU-based stages with 
visual-only criteria yields similar accuracy ($29.4\%$) but only marginally 
reduces latency ($0.92$s), since visual feature extraction dominates the 
cost.} Our cascaded MV filtering uses lightweight egomotion checks to
discard redundancy before applying visual criteria, achieving comparable
accuracy ($29.8\%/60.2\%$) at $0.61$s.

\noindent\textbf{Effectiveness of asymmetric cross-modal fusion.} 
The Concat+MLP method exhibits only $19.2\%$ on ScanQA, indicating that 
direct concatenation fails to correlate motion with visual content. 
Single-layer cross-attention improves accuracy to $26.1\%$ but remains 
suboptimal. Our asymmetric design achieves the best accuracy 
($29.8\%/60.2\%$) with only $+0.03$s additional latency as it allows 
motion tokens to learn visual context before passing information back to 
the visual stream. \blue{Notably, the Concat+MLP baseline is 
parameter-matched to our asymmetric design (95M vs.~110M) yet 
underperforms by $-10.6\%$ on ScanQA, confirming that the gain is 
primarily architectural.}


\vspace*{-0.05in}
\section{Conclusion}
\label{sec:conclusion}
\vspace*{-0.05in}

We proposed \sysname, a framework that enhances MLLMs with 
egomotion data from IMU sensors for 3D scene understanding. 
\sysname introduces two key components: a cascaded motion-visual 
keyframe filtering module that efficiently selects informative 
frames by progressing from lightweight IMU-based checks to 
demanding visual analysis, and an asymmetric cross-modal fusion 
module where motion tokens channel egomotion cues and cross-frame 
visual context into the visual representation. Experiments on 
five benchmarks demonstrate that \sysname significantly 
outperforms 2D-input baselines and exhibits accuracy comparable
to 3D-input methods \blue{while running $1.30\times$ and 
$1.61\times$ faster at matched accuracy, respectively}.

\noindent\textbf{Limitations and Future Work.}
\blue{Our evaluation focuses on indoor scenes with smooth handheld motion, 
with real IMU validation only on six TUM-VI sequences, and assumes 
hardware-synchronized video and IMU streams. Future work will extend the 
framework to outdoor, vehicle-mounted, and dynamic scenarios, validate 
on broader IMU hardware, and address unsynchronized sensor settings.}





{
\small
\bibliographystyle{plainnat}
\bibliography{reference}
}


\appendix

\section*{Technical Appendices and Supplementary Material}

\section{Additional Method Details}
\label{appendix:method}

\subsection{Details about Cascaded Motion-Visual Keyframe Filtering}
\label{appendix:keyframe-filtering}

Our keyframe filtering module applies three cascaded stages to each candidate frame $f_t$, evaluated against the most recently selected keyframe $\hat{f}_j$. Each stage acts as a progressively more expensive filter: (1) Stage1 uses IMU-derived motion estimates; (2) Stage2 uses sparse feature tracking; and (3) Stage3 uses visual token comparison. Below are the implementation details for each stage.

\noindent\textbf{Stage 1: Motion Gate.}
The translational displacement between the candidate frame $f_t$ and the last keyframe $\hat{f}_j$ is obtained by double-integrating gravity-corrected accelerometer readings:
\begin{equation}
d(\hat{f}_j, f_t) = \left\| \Delta\mathbf{p}_{j \to t} \right\|_2,
\end{equation}
where $\Delta\mathbf{p}_{j \to t}$ is the position change estimated from double integration. \blue{The velocity state is maintained continuously across the entire sequence, so constant-velocity segments (where the accelerometer reads near-zero after gravity correction) produce non-zero displacement.} The rotation angle is obtained by first integrating the gyroscope readings to form a relative rotation matrix $\Delta\mathbf{R}_{j \to t}$, then extracting the angle:
\begin{equation}
\theta(\hat{f}_j, f_t) = \arccos\!\left(\frac{\mathrm{tr}(\Delta\mathbf{R}_{j \to t}) - 1}{2}\right) \in [0, \pi].
\end{equation}
A candidate frame $f_t$ is discarded if both conditions hold: $d(\hat{f}_j, f_t) < \tau_d$ and $\theta(\hat{f}_j, f_t) < \tau_\theta$. We set $\tau_d = 0.2\,\text{m}$ and $\tau_\theta = 15^{\circ}$ based on the geometric properties of typical indoor scenes. At indoor object distances ($1\sim5$ m), a translation below 0.2 m produces insufficient parallax for meaningful geometric change. Similarly, a rotation below $15^{\circ}$ introduces less than 27\% new field of view given the camera's typical horizontal FOV (${\sim}55^{\circ}$).

\noindent\textbf{Stage 2: Lightweight Geometric Change Detection.}
This stage follows the sparse feature tracking approach used in visual-inertial odometry systems~\cite{qin2018vins,mourikis2007multi}. Let $\{p_k^j\}_{k=1}^{K}$ be $K$ sparse feature points detected in the last keyframe $\hat{f}_j$. These points are tracked to the candidate frame $f_t$ using a widely deployed feature tracker~\cite{campos2021orb}, yielding matched points $\{p_k^t\}$. The average parallax is computed as:
\begin{equation}
\bar{p}(\hat{f}_j, f_t) = \frac{1}{K'} \sum_{k=1}^{K'} \left\|p_k^t - p_k^j\right\|_2,
\end{equation}
where $K'$ is the number of successfully tracked points. A candidate is discarded if $\bar{p}(\hat{f}_j, f_t) < \tau_p$, with $\tau_p = 15$ pixels. This is a conservative threshold that retains candidates with moderate geometric change for further evaluation in Stage~3.

\noindent\textbf{Stage 3: Visual Token Analysis.}
Following the dual-encoder architecture in Spatial-MLLM~\cite{wu2025spatial}, the 2D encoder (Qwen2.5-VL's visual encoder~\cite{bai2025qwen25vltechnicalreport}) produces features $\mathbf{v}_{\mathrm{2D}}$, and the spatial encoder (VGGT backbone~\cite{wang2025vggt}) produces features $\mathbf{v}_{\mathrm{3D}}$. The 3D features $\mathbf{v}_{\mathrm{3D}}$ are rearranged to align with $\mathbf{v}_{\mathrm{2D}}$ in spatial and temporal dimensions, yielding $\mathbf{v}'_{\mathrm{3D}}$. The fused visual token for candidate frame $f_t$ is:
\begin{equation}
\mathbf{v}_t = \mathrm{MLP}_{\mathrm{2D}}(\mathbf{v}_{\mathrm{2D}}) + \mathrm{MLP}_{\mathrm{3D}}(\mathbf{v}'_{\mathrm{3D}}).
\end{equation}
\blue{The cosine distance is computed between globally average-pooled $\mathbf{v}_t$ and $\mathbf{v}_j$, determining whether $f_t$ is selected as a new keyframe. The unpooled spatial tokens are retained for downstream fusion.} Specifically, $f_t$ is selected if the cosine distance exceeds $\tau_v = 0.4$ (corresponding to a cosine similarity of $0.6$). Since $\mathbf{v}_t$ encodes both 2D appearance and 3D geometric structure, the cosine distance captures changes in both modalities. This threshold is tuned on the ScanQA~\cite{azuma2022scanqa} and SQA3D~\cite{ma2022sqa3d} datasets.

\begin{algorithm}[t]
\caption{Cascaded Motion-Visual Keyframe Filtering}
\label{alg:filtering}
\begin{algorithmic}[1]
\REQUIRE Video frames $\{f_1, \ldots, f_T\}$, IMU data, thresholds $\tau_d$, $\tau_\theta$, $\tau_p$, $\tau_v$
\ENSURE Keyframe set $\mathcal{K}$, visual tokens $\mathbf{V}$
\STATE $\mathcal{K} \leftarrow \{f_1\}$ ; $\hat{f}_j \leftarrow f_1$ ; extract $\mathbf{v}_j$ for $f_1$ ; $\mathbf{V} \leftarrow \{\mathbf{v}_j\}$
\FOR{$t = 2$ \TO $T$}
\STATE Compute $d(\hat{f}_j, f_t)$ and $\theta(\hat{f}_j, f_t)$ from IMU
\IF{$d(\hat{f}_j, f_t) < \tau_d$ \AND $\theta(\hat{f}_j, f_t) < \tau_\theta$}
\STATE \textbf{continue} \COMMENT{Stage 1: insufficient motion}
\ENDIF
\STATE Compute average parallax $\bar{p}(\hat{f}_j, f_t)$
\IF{$\bar{p}(\hat{f}_j, f_t) < \tau_p$}
\STATE \textbf{continue} \COMMENT{Stage 2: insufficient geometric change}
\ENDIF
\STATE Extract visual token $\mathbf{v}_t$
\IF{$1 - \cos(\mathbf{v}_t, \mathbf{v}_j) < \tau_v$}
\STATE \textbf{continue} \COMMENT{Stage 3: visually similar}
\ENDIF
\STATE $\mathcal{K} \leftarrow \mathcal{K} \cup \{f_t\}$ ; $\hat{f}_j \leftarrow f_t$ ; $\mathbf{v}_j \leftarrow \mathbf{v}_t$ ; $\mathbf{V} \leftarrow \mathbf{V} \cup \{\mathbf{v}_t\}$
\ENDFOR
\end{algorithmic}
\end{algorithm}

\noindent\textbf{Algorithm and Filtering Statistics.}
The complete cascaded filtering procedure is summarized in Algorithm~\ref{alg:filtering}. On our experimental benchmarks (ScanQA~\cite{azuma2022scanqa}, SQA3D~\cite{ma2022sqa3d}, VSI-Bench~\cite{yang2025thinking}, ScanRefer~\cite{chen2020scanrefer}, and Scan2Cap~\cite{chen2021scan2cap}, mostly derived from ScanNet~\cite{dai2017scannet}), input videos contain on average ${\sim}1{,}000$ frames at 30 FPS. Stage1 retains ${\sim}8\%$ (${\sim}80$ frames), Stage2 retains ${\sim}38\%$ of those (${\sim}30$ frames), and Stage3 retains ${\sim}70\%$ of those (${\sim}21$ keyframes). The cascaded design ensures that expensive visual feature extraction in Stage3 is applied to less than $3\%$ of the original frames.

\blue{\noindent\textbf{Threshold Sensitivity.} We additionally perturb each of the four thresholds ($\tau_d$, $\tau_\theta$, $\tau_p$, $\tau_v$) independently by $\pm 50\%$ from its default value and re-run the keyframe filtering pipeline on ScanQA. Across all eight perturbations, the resulting ScanQA EM changes by at most $1.2\%$ and the number of selected keyframes changes by at most $14.3\%$. This indicates that the cascaded filtering pipeline is robust to moderate threshold variations. We use the same threshold configuration ($\tau_d=0.2$\,m, $\tau_\theta=15^{\circ}$, $\tau_p=15$\,px, $\tau_v=0.4$) across all five benchmarks without per-dataset tuning, supporting cross-dataset generalization.}

\subsection{Egomotion Data Synthesis}
\label{appendix:egomotion-data}

Existing 3D scene understanding and spatial reasoning benchmarks lack raw IMU sensor data, as they were originally designed for vision-only or point-cloud-based methods. To enable training and evaluation with egomotion input, we synthesize realistic 6-axis IMU measurements (3-axis accelerometer and 3-axis gyroscope) \blue{from the per-frame camera pose annotations released by the source datasets}. \blue{For each scan, we define a world coordinate frame whose origin coincides with the initial camera position at $t_0$ and whose Z-axis is anti-parallel to gravity. All poses and measurements are expressed relative to this frame, allowing the gravity vector to take the canonical form $\mathbf{g} = [0, 0, -9.81]^\top$ m/s$^2$. The initial camera's orientation $\mathbf{R}_0$ relative to this frame is derived from the dataset's gravity calibration.}

\blue{\noindent\textbf{Source Pose Origins.} The four ScanNet-derived benchmarks (ScanQA, SQA3D, ScanRefer, Scan2Cap) provide per-frame camera-to-world transformation matrices $\mathbf{T}_i \in SE(3)$ at 30~FPS, which we use directly. VSI-Bench is sourced from the validation splits of three datasets and provides per-video metadata that links each clip back to its source scan~\cite{yang2025thinking}, from which we retrieve the corresponding ground-truth poses: ScanNet poses come from offline RGB-D bundle adjustment~\cite{dai2017scannet}, ScanNet++ poses from ARKit visual-inertial tracking with offline mesh alignment~\cite{yeshwanth2023scannet++}, and ARKitScenes poses from ARKit fused visual-inertial tracking~\cite{baruch2021arkitscenes}. All three release at metric scale and are routinely used as ground truth in 3D scene understanding tasks. The same B-spline differentiation procedure described below is applied across all five benchmarks.}

\noindent\textbf{\blue{B-spline Differentiation.}}
\blue{Given the per-scan pose trajectory described above,} we fit cubic B-splines~\cite{lovegrove2013spline} to the discrete position trajectory $\mathbf{p}(t) \in \mathbb{R}^3$ and rotation trajectory $\mathbf{R}(t) \in SO(3)$ separately, yielding a continuous-time representation that allows analytical computation of derivatives at arbitrary time instants. The noise-free accelerometer and gyroscope readings in the body frame are:
\begin{equation}
\tilde{\mathbf{a}}(t) = \mathbf{R}(t)^\top \left(\ddot{\mathbf{p}}(t) - \mathbf{g}\right), \qquad
\tilde{\boldsymbol{\omega}}(t) = \left(\mathbf{R}(t)^\top \dot{\mathbf{R}}(t)\right)^\vee,
\end{equation}
where $\ddot{\mathbf{p}}(t)$ is the linear acceleration in the world frame obtained from the second derivative of the position spline, and $(\cdot)^\vee$ extracts the angular velocity vector from the skew-symmetric matrix $\mathbf{R}^\top \dot{\mathbf{R}}$. \blue{The $-\mathbf{g}$ and $\mathbf{R}(t)^\top$ terms explicitly apply gravity subtraction and the world-to-body rotation, so that $\tilde{\mathbf{a}}(t)$ and $\tilde{\boldsymbol{\omega}}(t)$ match the body-frame format of real MEMS IMU sensors.}

\begin{table}[t!]
\centering
\caption{\textbf{IMU noise model parameters used in egomotion data synthesis.}}
\label{tab:imu-noise}
\resizebox{.65\textwidth}{!}{%
\begin{tabular}{@{}lcc@{}}
\toprule
\textbf{Parameter} & \textbf{Symbol} & \textbf{Value} \\
\midrule
Accelerometer noise density & $\sigma_a$ & $2.0 \times 10^{-2}$ m/s$^2$/$\sqrt{\text{Hz}}$ \\
Gyroscope noise density & $\sigma_g$ & $1.5 \times 10^{-3}$ rad/s/$\sqrt{\text{Hz}}$ \\
Accelerometer bias random walk & $\sigma_{ba}$ & $3.0 \times 10^{-3}$ m/s$^3$/$\sqrt{\text{Hz}}$ \\
Gyroscope bias random walk & $\sigma_{bg}$ & $2.0 \times 10^{-5}$ rad/s$^2$/$\sqrt{\text{Hz}}$ \\
\bottomrule
\end{tabular}%
}
\end{table}
\noindent\textbf{Noise Model and Sampling Rate.}
To simulate realistic sensor characteristics, we augment the ideal readings with additive white noise and slowly drifting bias, following the standard continuous-time IMU noise model~\cite{forster2016manifold}:
\begin{equation}
\begin{aligned}
\mathbf{a}(t) &= \tilde{\mathbf{a}}(t) + \mathbf{b}_a(t) + \mathbf{n}_a, \quad &\mathbf{n}_a &\sim \mathcal{N}(\mathbf{0},\, \sigma_a^2 \mathbf{I}) \\
\boldsymbol{\omega}(t) &= \tilde{\boldsymbol{\omega}}(t) + \mathbf{b}_g(t) + \mathbf{n}_g, \quad &\mathbf{n}_g &\sim \mathcal{N}(\mathbf{0},\, \sigma_g^2 \mathbf{I})
\end{aligned}
\end{equation}
where the biases $\mathbf{b}_a(t)$ and $\mathbf{b}_g(t)$ evolve as Brownian motions with $\dot{\mathbf{b}}_a \sim \mathcal{N}(\mathbf{0},\, \sigma_{ba}^2 \mathbf{I})$ and $\dot{\mathbf{b}}_g \sim \mathcal{N}(\mathbf{0},\, \sigma_{bg}^2 \mathbf{I})$. We use noise parameters representative of a consumer-grade MEMS IMU, as listed in Tab.~\ref{tab:imu-noise}. All IMU data is synthesized at 200\,Hz, matching common consumer sensor sampling rates~\cite{burri2016euroc}.


\subsection{Training Data Statistics}
\label{appendix:training-data}

We train \sysname on a mixture of four datasets, including ScanQA~\cite{azuma2022scanqa}, SQA3D~\cite{ma2022sqa3d}, ScanRefer~\cite{chen2020scanrefer}, and Scan2Cap~\cite{chen2021scan2cap}, all derived from ScanNet~\cite{dai2017scannet} indoor scans. Tab.~\ref{tab:data-stats} summarizes the dataset composition. Following~\cite{zheng2025video}, each training batch is randomly sampled from a single task type. VSI-Bench~\cite{yang2025thinking} is used only for evaluation, as it draws from additional data sources (ScanNet++~\cite{yeshwanth2023scannet++} and ARKitScenes~\cite{baruch2021arkitscenes}) and lacks a public training split.

\begin{table}[t!]
\centering
\caption{\textbf{Training and evaluation data statistics.} All training sets are derived from ScanNet~\cite{dai2017scannet}. $^\dagger$Scan2Cap reuses ScanRefer annotations with a different task formulation.}
\label{tab:data-stats}
\resizebox{0.6\textwidth}{!}{%
\begin{tabular}{@{}lcccc@{}}
\toprule
\textbf{Dataset} & \textbf{Task} & \textbf{\#Train} & \textbf{\#Eval} & \textbf{\begin{tabular}[c]{@{}c@{}}\#Scenes \\ (train)\end{tabular}} \\
\midrule
ScanQA~\cite{azuma2022scanqa} & 3D QA & 25{,}563 & 4{,}675 & 562 \\
SQA3D~\cite{ma2022sqa3d} & Situated QA & 26{,}623 & 3{,}519 & 518 \\
ScanRefer~\cite{chen2020scanrefer} & Visual Grounding & 36{,}665 & 9{,}508 & 562 \\
Scan2Cap$^\dagger$~\cite{chen2021scan2cap} & Dense Captioning & 36{,}665 & 9{,}508 & 562 \\
\midrule
\textbf{Total (training)} & & \textbf{125{,}516} & & \\
\midrule
VSI-Bench~\cite{yang2025thinking} & Spatial Reasoning & - & 5{,}131 & - \\
\bottomrule
\end{tabular}%
}
\end{table}

\subsection{Real IMU Benchmark Construction}
\label{appendix:tumvi-details}

\blue{\noindent\textbf{Source Data.} We use the six indoor room sequences of TUM-VI~\cite{schubert2018tum}, which provides hardware-synchronized fisheye video and 200\,Hz BMI160 IMU recordings. We undistort the fisheye images to pinhole-equivalent views using the calibration parameters provided by TUM-VI, producing a center-cropped pinhole video at approximately 80$^\circ$ effective field of view. The same undistortion is applied uniformly to all evaluated methods, removing fisheye distortion as a confounder.}

\blue{\noindent\textbf{Task Selection.} Among VSI-Bench's eight tasks, we cover six (Obj. Cnt., Abs. Dist., Room Size, Rel. Dist., Rel. Dir., Appr. Order) and exclude two: Obj. Size, whose reliable manual annotation requires 3D bounding-box annotations not provided by TUM-VI, and Route Planning, whose route definitions are difficult to standardize across short room sequences. Question availability per (sequence, task) cell varies with scene content: tasks with multiple natural object pairs (Rel. Dist., Rel. Dir., Appr. Order) admit more questions per room, while Room Size has limited natural variation (one ground-truth dimension per room). We allocate 24 questions each for Obj. Cnt.\ and Abs. Dist.\ (4 per room), 12 for Room Size (2 per room), and 20 each for the three multiple-choice tasks, totaling 120 QA pairs.}

\blue{\noindent\textbf{Annotation Methodology.} We follow the publicly released question templates of VSI-Bench~\cite{yang2025thinking}. Question candidates are generated by filling templates with object pairs identified from each sequence, with optional vision-language model assistance for variety. Ground-truth answers are determined from TUM-VI's mocap-tracked camera trajectory (for distance and dimension queries) and from manual video inspection (for counting, direction, and ordering queries). All annotations are manually reviewed, and a second annotator independently verifies a $20\%$ random subset to confirm reliability.}

\subsection {Prompts of \sysname}
\label{appendix:prompt}

\begin{figure*}[t!]
  \begin{center}
    \centerline{\includegraphics[width=\columnwidth]{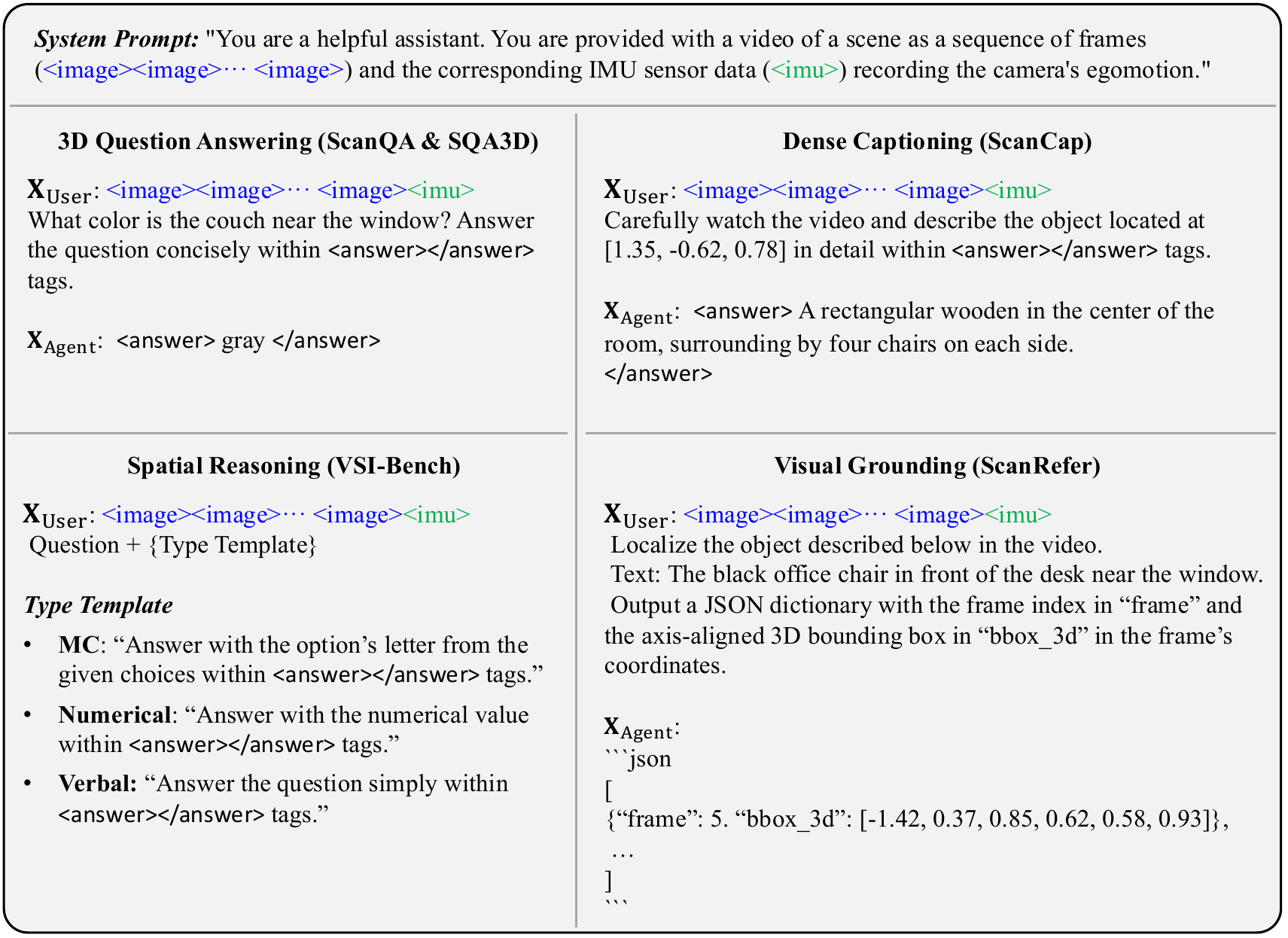}}
    \vspace*{-0.05in}
    \caption{
      \textbf{Prompts used for each task in \sysname.} All tasks share the same system prompt and receive video frames along with IMU data as input. Each task uses a task-specific user prompt with a unified \texttt{<answer>} tag format for response extraction.
    }
    \label{fig:prompts}
  \end{center}
  \vspace*{-0.5in}
\end{figure*}

Fig.~\ref{fig:prompts} shows the prompts used in \sysname for each task. We adopt the default system prompt of Qwen2.5-VL~\cite{bai2025qwen25vltechnicalreport}, ``You are a helpful assistant,'' for all tasks. The user prompt contains video frames and IMU data as multimodal input, followed by a task-specific text instruction. All tasks use a unified \texttt{<answer>} tag format for response extraction.

For question answering (ScanQA~\cite{azuma2022scanqa}, SQA3D~\cite{ma2022sqa3d}, and VSI-Bench~\cite{yang2025thinking}), the model is instructed to answer concisely. For SQA3D, the situation description is prepended to the question. For VSI-Bench, the prompt appends a type-specific template depending on the question type (multiple choice, numerical, or verbal). For visual grounding (ScanRefer~\cite{chen2020scanrefer}), the model outputs a JSON dictionary containing the frame index and an axis-aligned 3D bounding box \blue{in the predicted frame's camera coordinates. For IoU computation, the predicted box corners are transformed to the world frame via the camera-to-world pose of the predicted frame and re-axis-aligned, then compared against the world-frame ground-truth}. For dense captioning (Scan2Cap~\cite{chen2021scan2cap}), the model describes the object at a given 3D location.

\section{Additional Evaluation Results}

\subsection{Full Quantitative Results on ScanQA and SQA3D}
\label{appendix:scanqa-sqa3d}

\begin{table}[t!]
\centering
\caption{\textbf{Full evaluation Results on ScanQA~\cite{azuma2022scanqa}.} We use the val set of ScanQA for evaluation. Reported metrics include EM-1, BLEU-1 to BLEU-4, METEOR, ROUGE-L, and CIDEr. "–" indicates the number is not available. ``2D'', ``3D'', and ``M'' specify the model's input type as 2D data (images/videos), 3D data (point clouds/depth maps), and egomotion data, respectively.}
\label{tab:appendix_ScanQA}
\resizebox{\textwidth}{!}{%
\begin{tabular}{@{}l|ccc|cccccccc@{}}
\toprule
\multirow{2}{*}{\textbf{Methods}} & \multicolumn{3}{c|}{\textbf{Input}} & \multicolumn{8}{c}{\textbf{ScanQA (val)}} \\ \cmidrule(l){2-12}
                                  & 2D          & 3D        & M         & EM-1          & BLEU-1        & BLEU-2        & BLEU-3        & BLEU-4        & METEOR        & ROUGE-L       & CIDEr          \\ \midrule
\textit{Task-Specific Models}     &             &           &           &               &               &               &               &               &               &               &                \\
ScanQA~\cite{azuma2022scanqa}                            &            & \Checkmark &           & 21.1          & \textbf{30.2}          & \textbf{20.4} & \textbf{15.1} & 10.1          & 13.1          & 33.3          & 64.9           \\
3D-VisTA~\cite{zhu20233d}                          &            & \Checkmark &           & \textbf{22.4} & -             & -             & -             & \textbf{10.4}          & \textbf{13.9} & \textbf{35.7} & \textbf{69.6}  \\ \midrule
\textit{2D-Input Models}       &             &           &           &               &               &               &               &               &               &               &                \\
Qwen2.5-VL-3B~\cite{bai2025qwen25vltechnicalreport}                     & \Checkmark  &           &           & 15.4          & 22.5          & 13.1          & 8.1           & 3.8           & 9.7           & 25.4          & 47.4           \\
Qwen2.5-VL-7B~\cite{bai2025qwen25vltechnicalreport}                      & \Checkmark  &           &           & 19.0          & 27.8          & 13.6          & 6.3           & 3.0           & 11.4          & 29.3          & 53.9           \\
LLaVA-Video-7B~\cite{zhang2024llava}                    & \Checkmark  &           &           & -             & 39.7          & 26.6          & 9.3           & 3.1           & 17.7          & 44.6          & 88.7           \\
VideoChat2~\cite{li2025videochat}                        & \Checkmark  &           &           & 19.2          & -             & -             & -             & 9.6           & 9.5           & 28.2          & 49.2           \\
Spatial-MLLM~\cite{wu2025spatial}                      & \Checkmark  &           &           & \textbf{26.3} & \textbf{44.4} & \textbf{28.8} & \textbf{21.9} & \textbf{14.8} & \textbf{18.4} & \textbf{45.0} & \textbf{91.8}  \\  \midrule
\textit{3D/2.5D-Input Models}     &             &           &           &               &               &               &               &               &               &               &                \\
3D-LLM~\cite{hong20233d}                            & \Checkmark  & \Checkmark &           & 20.5          & 39.3          & 25.2          & 18.4          & 12.0          & 14.5          & 35.7          & 69.4           \\
Chat-Scene~\cite{huang2024chat}                         & \Checkmark  & \Checkmark &           & 21.6          & 43.2          & 29.1          & 20.6          & 14.3          & 18.0          & 41.6          & 87.7           \\
LEO~\cite{huang2023embodied}                               & \Checkmark  & \Checkmark &           & 24.5          & -             & -             & -             & 13.2          & 20.0          & 49.2          & 24.5           \\
GPT4Scene-HDM~\cite{qi2025gpt4scene}                               & \Checkmark  & \Checkmark &           & 28.2          & 44.4          & 30.3          & 22.3          & 15.5          & 18.9          & 46.5          & 96.3           \\
Video-3D LLM~\cite{zheng2025video}                      & \Checkmark  & \Checkmark &           & 30.1          & \textbf{47.1} & \textbf{31.7} & \textbf{22.8} & 16.2          & 19.8          & 49.0          & 102.1          \\
LLaVA-3D~\cite{zhu2024llava}                          & \Checkmark  & \Checkmark &           & \textbf{30.6} & -             & -             & -             & \textbf{16.4} & \textbf{20.8} & \textbf{49.6} & \textbf{103.1} \\ \midrule
\textit{Egomotion-Input Model}    &             &           &           &               &               &               &               &               &               &               &                \\
\textbf{\sysname-4B}                       & \Checkmark  &           & \Checkmark & \textbf{29.8} & \textbf{46.5} & \textbf{31.2} & \textbf{22.2} & \textbf{15.8} & \textbf{20.3} & \textbf{48.0} & \textbf{100.4} \\ \bottomrule
\end{tabular}%
}
\end{table}

\begin{table}[t!]
\centering
\caption{\textbf{Evaluation Results on SQA3D~\cite{ma2022sqa3d}.} We use the test set of SQA3D for evaluation. We additionally report the average EM@1 for different question types, including \textit{What}, \textit{Is}, \textit{How}, \textit{Can}, \textit{Which}, and \textit{Others}. "–" indicates the number is not available. ``2D'', ``3D'', and ``M'' specify the model's input type as 2D data (images/videos), 3D data (point clouds/depth maps), and egomotion data, respectively.}
\label{tab:appendix_SQA3D}
\resizebox{\textwidth}{!}{%
\begin{tabular}{@{}l|ccc|cccccc|cc@{}}
\toprule
\multirow{2}{*}{\textbf{Methods}} & \multicolumn{3}{c|}{\textbf{Input}} & \multicolumn{8}{c}{\textbf{SQA3D (test)}} \\ \cmidrule(l){2-12}
                                  & 2D          & 3D        & M         & What          & Is            & How           & Can           & Which         & Others        & Avg. (EM@1)          & Avg. (EM@R1)        \\ \midrule
\textit{Task-Specific Models}     &             &           &           &               &               &               &               &               &               &                      &                     \\
SQA3D~\cite{ma2022sqa3d}                             &            & \Checkmark &           & 31.6          & \textbf{63.8} & \textbf{46.0} & 69.5          & 43.9          & 45.3          & 46.6                 & -                   \\
3D-VisTA~\cite{zhu20233d}                          &            & \Checkmark &           & \textbf{34.8} & 63.3          & 45.4          & \textbf{69.8} & \textbf{47.2} & \textbf{48.1} & \textbf{48.5}        & -                   \\ \midrule
\textit{2D-Input Models}       &             &           &           &               &               &               &               &               &               &                      &                     \\
Qwen2.5-VL-3B~\cite{bai2025qwen25vltechnicalreport}                     & \Checkmark  &           &           & 34.8          & 52.1          & 39.8          & 52.7          & 45.6          & 47.0          & 43.4                 & 45.9                \\
Qwen2.5-VL-7B~\cite{bai2025qwen25vltechnicalreport}                      & \Checkmark  &           &           & 39.7          & 56.6          & 41.1          & 55.9          & 47.6          & 47.2          & 46.5                 & 49.8                \\
LLaVA-Video-7B~\cite{zhang2024llava}                    & \Checkmark  &           &           & 42.7          & 56.3          & 47.5          & 55.3          & 50.1          & 47.2          & 48.5                 & -                   \\
Spatial-MLLM~\cite{wu2025spatial}                      & \Checkmark  &           &           & \textbf{45.9} & \textbf{71.6} & \textbf{55.1} & \textbf{69.5} & \textbf{52.0} & \textbf{53.0} & \textbf{55.9}        & \textbf{58.7}       \\  \midrule
\textit{3D/2.5D-Input Models}     &             &           &           &               &               &               &               &               &               &                      &                     \\
Chat-Scene~\cite{huang2024chat}                         & \Checkmark  & \Checkmark &           & 45.4          & 67.0          & 52.0          & 69.5          & 49.9          & 55.0          & 54.6                 & 57.5                \\
GPT4Scene-HDM~\cite{qi2025gpt4scene}                               & \Checkmark  & \Checkmark &           & \textbf{55.9} & 69.9          & 50.8          & 68.7          & \textbf{53.3} & \textbf{60.4} & \textbf{60.6}        & \textbf{63.3}       \\
Video-3D LLM~\cite{zheng2025video}                      & \Checkmark  & \Checkmark &           & 51.1          & \textbf{72.4} & \textbf{55.5} & \textbf{69.8} & 51.3          & 56.0          & 58.6                 & -                   \\ \midrule
\textit{Egomotion-Input Model}    &             &           &           &               &               &               &               &               &               &                      &                     \\
\textbf{\sysname-4B}                       & \Checkmark  &           & \Checkmark & \textbf{54.1} & \textbf{72.0} & \textbf{54.8} & \textbf{70.3} & \textbf{53.0} & \textbf{58.9} & \textbf{60.2}        & \textbf{62.4}       \\ \bottomrule
\end{tabular}%
}
\end{table}

Tab.~\ref{tab:appendix_ScanQA} extends Tab.~\ref{tab:ScanQA} with the complete metric suite on ScanQA~\cite{azuma2022scanqa}, additionally reporting EM-1, BLEU-2, and BLEU-3. The results are fully consistent with the main findings: \sysname achieves the best EM-1 among 2D-input models ($29.8$ vs.\ $26.3$ for Spatial-MLLM) and remains comparable with top 3D-input models across all BLEU levels, confirming that the trends in Tab.~\ref{tab:ScanQA} hold across the full metric set.

Tab.~\ref{tab:appendix_SQA3D} provides a question-type breakdown on SQA3D~\cite{ma2022sqa3d}. \sysname outperforms all 2D-input models across nearly all question types, with the largest gains on spatially-grounded and situational categories (i.e., \textit{What}, \textit{Is}, \textit{Can}). Compared to 3D/2.5D-input models, \sysname achieves comparable or superior scores on most question types, demonstrating that lightweight egomotion data can match the spatial grounding capability of expensive 3D inputs. On \textit{How} questions, which are typically quantitative count tasks, \sysname slightly underperforms the SOTA 2D-input baseline (Spatial-MLLM, $-0.3$), suggesting that egomotion cues are less decisive when visual information dominates. These per-type results are consistent with the conclusion in Sec.~\ref{subsec:eval_spatial_reasoning}.

\blue{\noindent\textbf{Per-seed Variance.} Across Tabs.~\ref{tab:ScanQA}--\ref{tab:ablation}, 3-seed standard deviations remain within $\pm 0.5$ on EM/accuracy metrics and $\pm 1.5$ on caption metrics (CIDEr).}

\subsection{Pareto Analysis of Cost-Effectiveness}
\label{appendix:pareto}

\begin{figure}[t!]
  \centering
  \includegraphics[width=\textwidth]{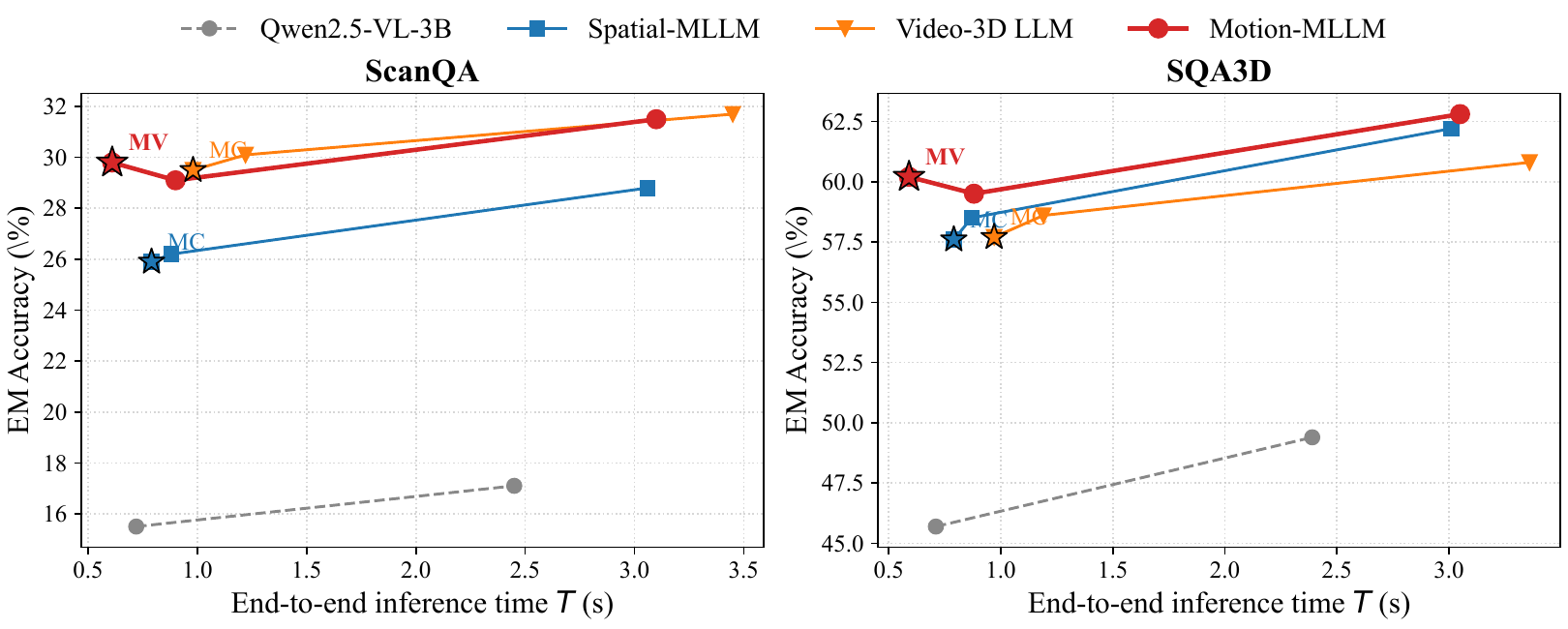}
  \vspace*{-0.2in}
  \caption{\textbf{Latency-accuracy trade-off on ScanQA (left) and SQA3D (right).} Each baseline polyline traces its frame-sampling configurations. \sysname uses MV-filtering.}
  \label{fig:pareto}
\end{figure}

\blue{The CE ratio used in Sec.~\ref{subsec:cost_effectiveness} is a single-number summary that implicitly weighs accuracy against latency. Fig.~\ref{fig:pareto} reports the full latency-accuracy trade-off, with each baseline shown as a polyline through its sampling configurations (uniform 128, uniform 32, and adaptive MC where applicable). \sysname's MV-filtering point sits at the low-latency end of the Pareto frontier on both benchmarks, strictly dominating every baseline at its best CE point. Baselines reach higher peak accuracy only at $\sim 5\times$ \sysname's latency, confirming that the efficiency advantage in Sec.~\ref{subsec:cost_effectiveness} is a genuine latency-accuracy property.}

\subsection{Qualitative Results}
\label{appendix:qualitative}

\begin{figure*}[t!]
  \begin{center}
    \centerline{\includegraphics[width=\columnwidth]{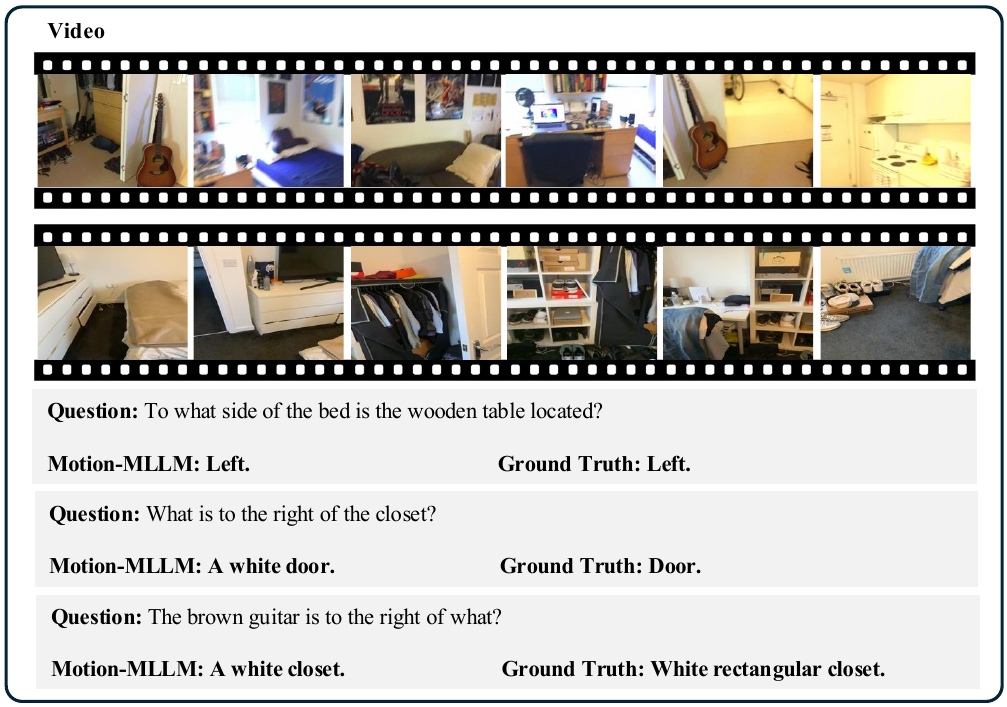}}
    \vspace*{-0.05in}
    \caption{
      \textbf{Qualitative examples on ScanQA~\cite{azuma2022scanqa}.}
    }
    \label{fig:scanqa_visualization}
  \end{center}
  \vspace*{-0.1in}
\end{figure*}

\begin{figure*}[t!]
  \begin{center}
    \centerline{\includegraphics[width=\columnwidth]{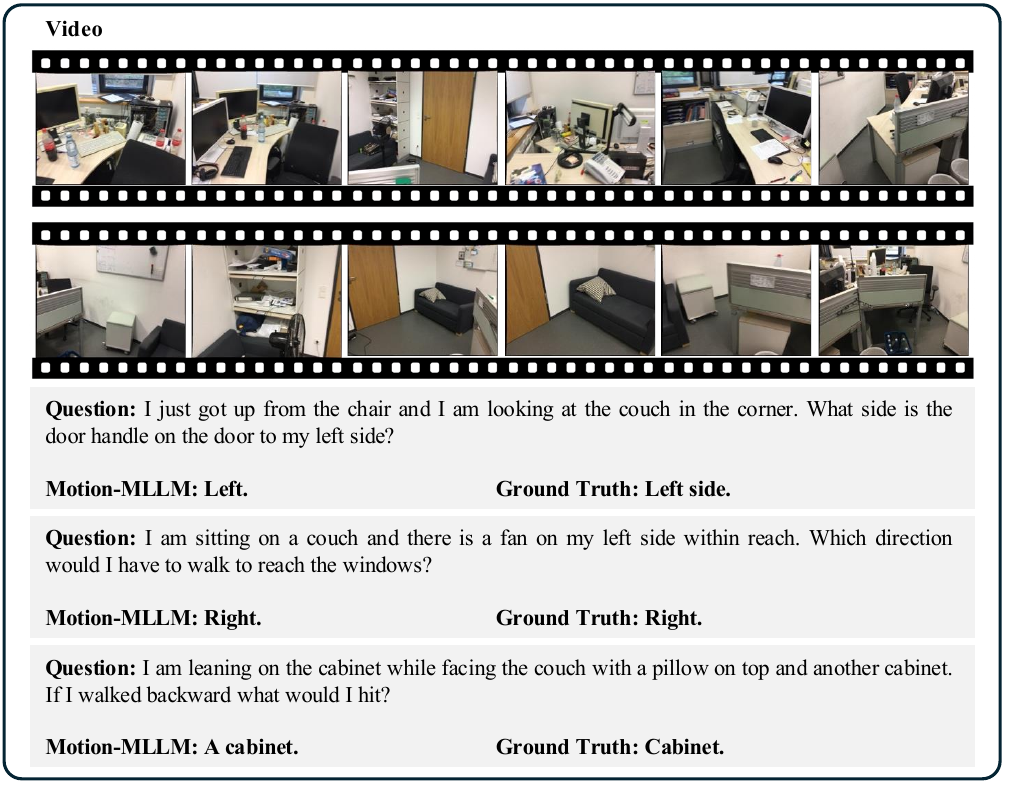}}
    \vspace*{-0.05in}
    \caption{
      \textbf{Qualitative examples on SQA3D~\cite{ma2022sqa3d}.}
    }
    \label{fig:sqa3d_visualization}
  \end{center}
  \vspace*{-0.1in}
\end{figure*}

\begin{figure*}[t!]
  \begin{center}
    \centerline{\includegraphics[width=\columnwidth]{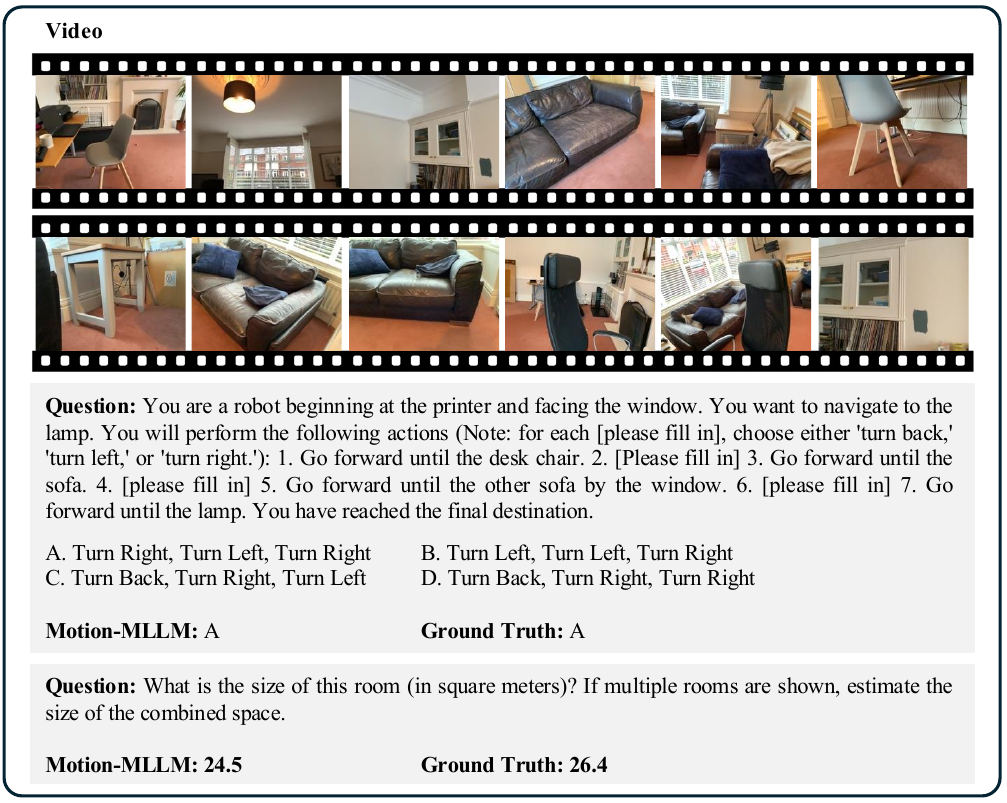}}
    \centerline{\includegraphics[width=\columnwidth]{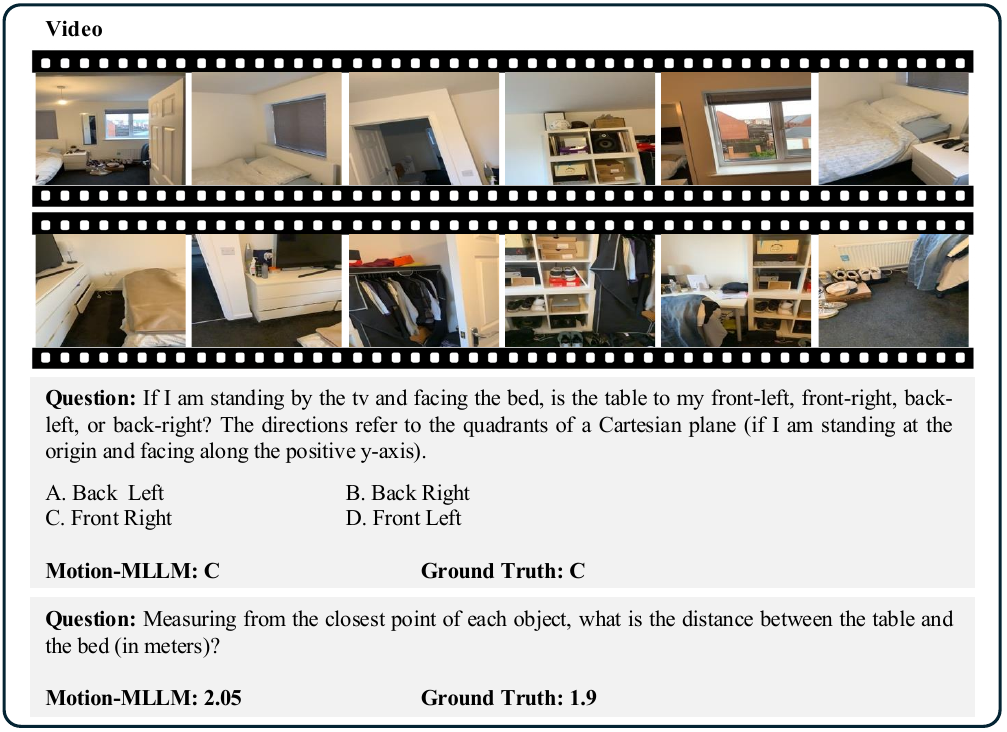}}
    \vspace*{-0.05in}
    \caption{
      \textbf{Qualitative examples on VSI-Bench~\cite{yang2025thinking}.}
    }
    \label{fig:vsi_visualization}
  \end{center}
  \vspace*{-0.3in}
\end{figure*}

\begin{figure*}[t!]
  \begin{center}
    \centerline{\includegraphics[width=\columnwidth]{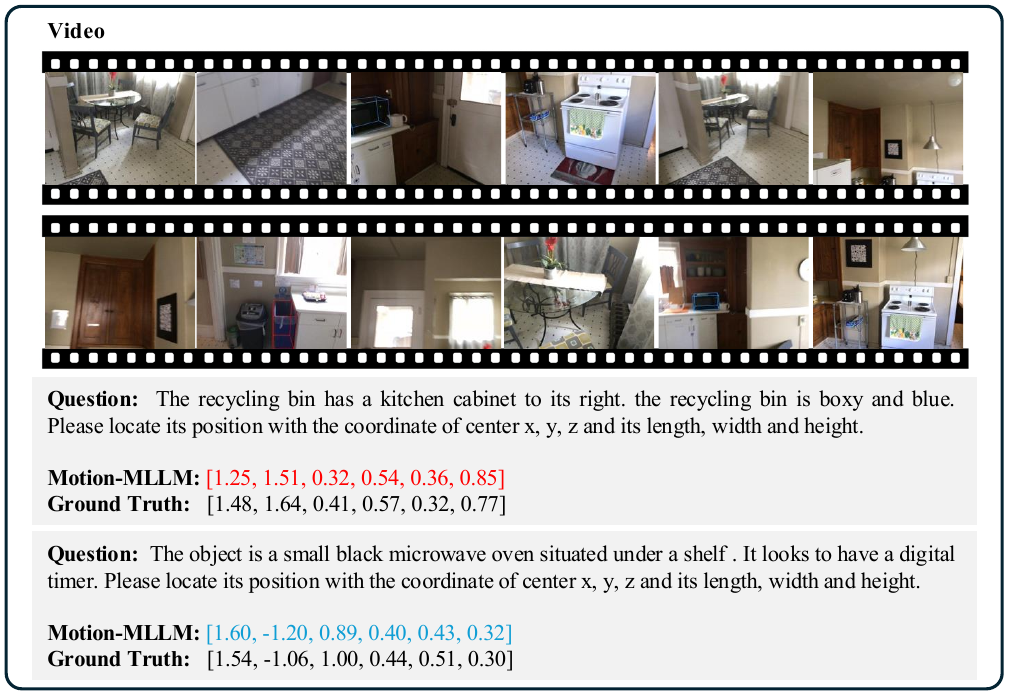}}
    \vspace*{-0.05in}
    \caption{
      \textbf{Qualitative examples of visual grounding on ScanRefer~\cite{chen2020scanrefer}.}
    }
    \label{fig:scanrefer_visualization}
  \end{center}
  \vspace*{-0.3in}
\end{figure*}

\begin{figure*}[t!]
  \begin{center}
    \centerline{\includegraphics[width=\columnwidth]{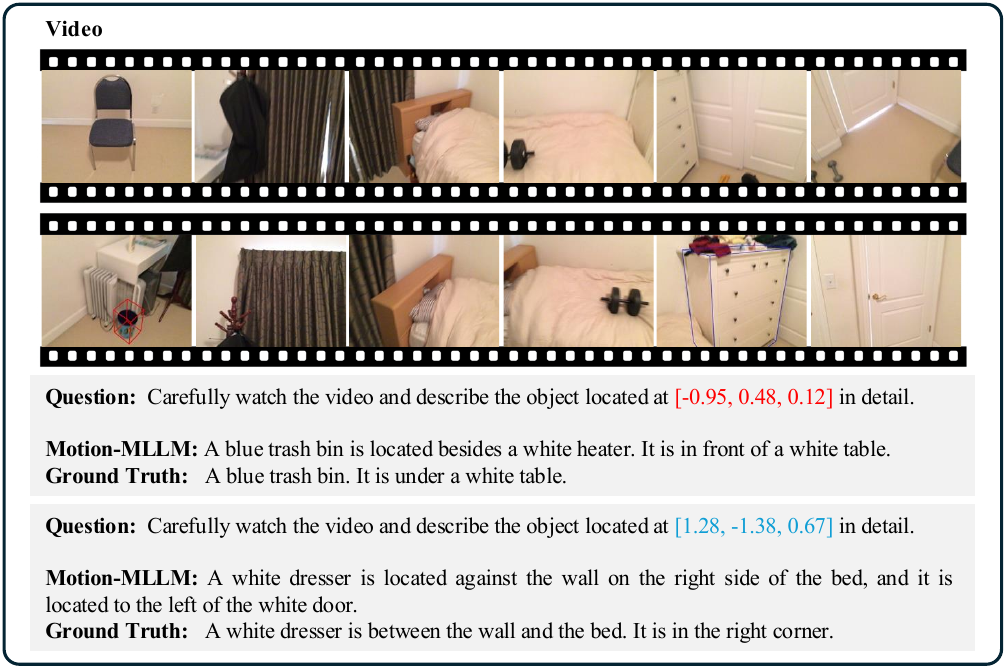}}
    \vspace*{-0.05in}
    \caption{
      \textbf{Qualitative examples of dense captioning on Scan2Cap~\cite{chen2021scan2cap}.}
    }
    \label{fig:scan2cap_visualization}
  \end{center}
  \vspace*{-0.3in}
\end{figure*}

We provide qualitative examples to visualize the evaluation results of \sysname on ScanQA~\cite{azuma2022scanqa} (Fig.\ref{fig:scanqa_visualization}), SQA3D~\cite{ma2022sqa3d} (Fig.\ref{fig:sqa3d_visualization}), VSI-Bench~\cite{yang2025thinking} (Fig.\ref{fig:vsi_visualization}), ScanRefer~\cite{chen2020scanrefer} (Fig.~\ref{fig:scanrefer_visualization}), and Scan2Cap~\cite{chen2021scan2cap} (Fig.~\ref{fig:scan2cap_visualization}). These examples show that \sysname correctly answers spatial questions, localizes described objects, and generates grounded captions across diverse indoor scenes.

\end{document}